\documentclass[10pt,a4paper]{article}
\usepackage[linesnumbered,boxruled,commentsnumbered]{algorithm2e}
\usepackage{inputenc}
\usepackage{graphicx}
\usepackage{amsmath}
\usepackage{caption}
\usepackage{latexsym}
\usepackage{morefloats}
\usepackage{floatrow}
\usepackage{multicol}
\usepackage{multirow}
\usepackage{float}
\usepackage{amsthm}
\floatstyle{plaintop}
 \usepackage{floatrow}
\restylefloat{table}
\usepackage{amsthm}
\usepackage{enumitem}
\usepackage{graphicx}
\usepackage{caption}
\usepackage{subcaption}

\makeatletter
\@addtoreset{proofpart}{thm}
\makeatother

\usepackage{epstopdf}
\begin{document}

\title{\textit{Fault Matters}: Sensor Data Fusion for Detection of Faults using Dempster-Shafer Theory of Evidence in IoT-Based Applications}
\author{Nimisha~Ghosh,
	Rourab~Paul,
	Satyabrata~Maity, Krishanu~Maity,\\
	Sayantan~Saha\\
	Institute of Technical Education and Research,\\
	 Siksha 'O' Anusandhan (Deemed to be University),\\Bhubaneswar 751030, Odisha, India
}
\maketitle	
\thispagestyle{empty}
\pagestyle{empty}	
		\begin{abstract}
		Fault detection in sensor nodes is a pertinent issue that has been an important area of research for a very long time. But it is not explored much as yet in the context of Internet of Things. Internet of Things work with a massive amount of data so the responsibility for guaranteeing the accuracy of the data also lies with it. Moreover, a lot of important and critical decisions are made based on these data, so ensuring its correctness and accuracy is also very important. Also, the detection needs to be as precise as possible to avoid negative alerts. For this purpose, this work has adopted Dempster-Shafer Theory of Evidence which is a popular learning method to collate the information from sensors to come up with a decision regarding the faulty status of a sensor node. To verify the validity of the proposed method, simulations have been performed on a benchmark data set and data collected through a test bed in a laboratory set-up. For the different types of faults, the proposed method shows very competent accuracy for both the benchmark (99.8\%) and laboratory data sets (99.9\%) when compared to the other state-of-the-art machine learning techniques.  
	\end{abstract}
	\textbf{\textit{Keywords}}: Classification, Data Fusion, Dempster-Shafer, Fault Detection, Internet of Things
%
\section{Introduction}
\label{intro}
Internet of Things (IoT) have extended computing capability to everyday objects which are embedded with sensors to come up with a huge repository of data which can be exchanged and consumed without any human intervention. This data plays a huge role in decision-making policies. Thus, the foremost requirement in any application of IoT would be to ensure the reliability of data. For this purpose, detection of a faulty device which produces anomalous data is very important. A data is considered to be faulty (anomalous or outlier) if it deviates significantly from the normal range of value~\cite{hawkins}. These deviations may turn out to be catastrophic if not detected on time. Faulty sensor data is attributed to the crash of Lion Air 737 on 29th October 2018 which killed all 189 people on-board. Erroneous signals from sensor caused continuous nose-down motions resulting in the ill-fated crash.

Outlier detection is a very pertinent topic in any field which is concerned with data like fraud and intrusion detection~\cite{HAJIHEIDARI2019}, weather monitoring~\cite{shepherd2003}, traffic anomaly detection~\cite{Xie}, sensor faults in heating, ventilation and air-conditioning (HVAC) systems~\cite{Reppa} etc. In an IoT application, monitoring and collection of real-time data from various sensors are necessary for the proper functioning of the system. All the data that is collected should be meticulously processed for the smooth operation of any system. The inherent components of any sensor network are prone to failures and this may result in incorrect readings. Like any other practical system, faulty sensor data may cause instability in IoT-based system as well. Thus, a proper fault detection technique needs to be incorporated to report the abnormality in the sensor data and thus identify a compromised node. Fault detection inherently can be considered to be a classification problem. Classification is a supervised learning technique in which labels are assigned to an observation based on its feature values. In a classification problem, data is first trained with a suitable learning technique so that the classifier learns the difference between the different classes. Later, the classifier can be used to test a new observation to make a conclusion about its class. This work uses Dempster-Shafer Theory of Evidence (DSTE)~\cite{Dempster2008,shafer197} which is a mathematical model that combines information from different sources. DSTE deals with uncertainty and thus can work with missing informations  and conflicting informations. Thus, it can be considered to be a more general approach than Bayesian theory of probability and has found its applicability in varied domains~\cite{ZERVAS,Xia,JAn,BQin,HZhu}. Faulty node detection which comprises of different kinds of faults as given in~\cite{MUHAMMED2017} can be yet another application where DSTE can be applied. 

The main contributions of this work can thus be given as follows:
\begin{itemize}
	\item Applying DSTE and devising a mass assignment function to create a classification method to detect different kinds of sensor faults in the context of IoT application.
	\item Verifying the applicability of the proposed method on the benchmark labelled data set~\cite{Suthaharan} and data collected from laboratory set up.
	\item The proposed method has also been compared with existing state-of-the-art classifiers to verify its feasibility.
\end{itemize}

The rest of the paper is organised as follows: Section~\ref{rls} gives the relevant literature study followed by Section~\ref{DST} which gives a preliminary idea about Dempster-Shafer theory of evidence. Next, in Section~\ref{fault}, the different types of faults pertaining to a sensor node is elaborated. In Section~\ref{pm}, the proposed method is discussed and in Section~\ref{ea}, the results are put forward based on the experimental analysis. Finally, Section~\ref{con} concludes the paper.
\section{Related Studies}
\label{rls}
There are many works which have addressed the existence of faults in sensor data but research in this context is further necessary due to exponential demand of ``smart" objects and relevant IoT technologies. Fault detection technique in any system can be divided into supervised, unsupervised and semi-supervised learning. In both supervised and semi-supervised learning works with labelled data to detect the fault, whereas unsupervised learning uses patterns to sort out the faults from the rest of the data. Different approaches like Hidden Markov models, distance, clustering are available for each of the learning techniques. The authors in~\cite{Warriach} proposed a technique to detect faults in sensors using Hidden Markov models where the experiment was carried out in real scenarios. In~\cite{Djenouri}, the authors used k-nearest neighbour method for distance-based outlier detection on spatio-temporal traffic flow. Cluster-based data analysis using recursive principal component analysis (R-PCA) is proposed in~\cite{Tyu} which aggregates redundant data from sensors and helps in detecting the outliers. Outlier mining through fuzzy set theory has been explored in~\cite{LJin}. Lu et.al~\cite{Wlu} used a deep structured framework for detection of outliers. Kernel principal component analysis based Mahalanobis kernel is yet another outlier detection method that has been applied in~\cite{Ghorbel}. In~\cite{Yessembayev}, the authors used Local Outlier Factor (LOF) algorithm to segregate normal nodes from anomalous ones. Based on density-based spatial clustering of applications with noise (DBSCAN), Abid et. al~\cite{Aabid} proposed an outlier detection technique for wireless sensor networks. An in-network knowledge discovery approach has been proposed in~\cite{FAWZY} for outlier detection in sensor nodes to differentiate between local outlier, cluster outlier and network outlier. Threshold-based fault detection and repairing scheme with Dynamic Bayesian Network (DBN) in intelligent connected vehicle (ICV) has been explored in~\cite{HZhang}. DBN can acquire the spatial and temporal correlations of vehicle data for precise fault detection. In~\cite{zidi2017}, the authors have used Support Vector Machines for identifying faults in sensor nodes using labelled data sets. Though there have been many works which have addressed fault or outlier detection in wireless sensor networks, not many have worked towards creating a framework for an IoT platform. This work intends to achieve the same by proposing a considerable framework for an IoT environment.
\section{Dempster-Shafer Theory of Evidence}
\label{DST}
Dempster-Shafer Theory of Evidence (DSTE) or the theory of belief functions is used to model and validate the uncertainty present in statistical deductions. The statical inference or deduction includes all the possible states of a system or in general the hypotheses. These hypotheses are then assigned probability assignments or mass functions and they are combined to reach a final decision. DSTE helps in fusion of sensor data by applying a combination rule on the mass functions of the data source. 

\subsection{Frame of Discernment}
It is the set of all hypotheses say, $\theta_1, \theta_2, \theta_3,\dots$  as represented by $\Theta$ where the elements are mutually exclusive and the set is exhaustive. Each subset of $\Theta$ can be considered to have a probable answer to a question. As the elements in $\Theta$ are mutually exclusive 
and the set is exhaustive, one only correct answer is possible. The set of all subsets is given by the power set $2^{\Theta}$. If there are three hypotheses $\theta_1, \theta_2, \theta_3$, then,
$$\Theta = \{\theta_1, \theta_2, \theta_3\}$$ and,
$$2^{\Theta} = \{\phi, \{\theta_1\}, \{\theta_2\}, \{\theta_3\}, \{\theta_1\theta_2\}, \{\theta_2\theta_3\}, \{\theta_1\theta_3\}, \{\theta_1\theta_2\theta_3\}\}$$

\subsection{Mass Assignment Function}
A \textbf{\textit{M}}ass \textbf{\textit{A}}ssignment \textbf{\textit{F}}unction (MAF) assigns each hypotheses $\theta_1, \theta_2, \theta_3,\dots, \theta_n$ to a mass value $m(\theta)$ in the range [0,1] with the conditions that:
\begin{enumerate}
	\item The mass value of an empty set is zero that is, $m(\phi) = 0$
	\item The summation of all possible hypotheses and their combinations must be equal to 1, that is, $\sum_{\theta\in 2^\Theta}m(\theta) = 1$
\end{enumerate}
\subsection{Dempster's Combination Rule}
Once individual mass assignments are done with, it becomes imperative to have a cumulative value in order to reach a conclusive decision based on sensor fusion. This problem is solved by applying Dempster's rule which provides the tool to combine mass assignments from multiple information sources. When two mass assignments are combined they can either produce a null set or they may have an intersection point. In the former case, the mass assignment is considered to have a zero value and the mass assignment of the non-empty set is boosted by the factor $K$, commonly known as the conflict factor, such that the summation of the non-empty set is equal to 1. Considering these factors, Dempster's rule to combine the mass assignment values from various sources is given by:
\begin{equation}
\label{eq:1}
m_1\oplus m_2 (Z) =\frac{\sum_{X \cap Y=Z\neq \phi} m_1(X)m_2(Y)}{1-K}
\end{equation}
Here, $K = \sum_{X \cap Y=\phi} m_1(X)m_2(Y)$ and $X,Y,Z \subseteq \Theta$. $\oplus$ is considered to be the orthogonal or direct sum. So, $m_1\oplus m_2 (Z)$ is the combined belief of two separate mass assignments and $\phi$ denotes the null set. The numerator in (\ref{eq:1}) encompasses all the possibilities whose intersection is $X \cap Y=Z$. To normalise this value, it is divided by $1-K$ which represents all the combined values that produces a null set. Dempster's rule of combination is iteratively applied on all the information sources to produce the final result.

\section{Types of Faults in IoT-Based Applications}
\label{fault}
Faults can occur in any application where sensor plays a major role in data collection. The collected data is usually represented in the form of a time series $d(n,t,f(t))$ where $f(t)$ is the value of the node $n$ at time $t$. $f(t)$ can be given by: $f(t) = \alpha + \beta x+\eta$, where $\alpha$ is called the offset, $\beta$ is the gain factor, $x$ is the non-faulty data and $\eta$ represents the noise in the data at time instance $t$. The various types of faults are given as below~\cite{MUHAMMED2017}:
\begin{enumerate}
	\item \textit{\textbf{Gain Fault}}: Gain faults take place if, over a period of time, rate of change of sensed data does not match with the expected data. It can be expressed as:
	\begin{equation}
	x' = \beta x + \eta 
	\label{eq:2}
	\end{equation}
	where, $x'$ is the faulty reading.
	\item \textit{\textbf{Offset Fault}}: Offset faults can be attributed to the addition of an additive constant to the sensed data. It can be represented as:
	\begin{equation}
	x' = \alpha + x + \eta 
	\label{eq:3}
	\end{equation}
	\item \textit{\textbf{Data-Loss Fault}}: Data loss fault occurs when a node has missing data during a particular time series. In this case, $f(t)$ has a null value.
	\item \textit{\textbf{Out-of-Bounds Fault}}: If the normal sensed value of data lies between say, $\gamma_1$ and $\gamma_2$, then out of bounds fault will occur for $x'\in f(t)$ if $x'< \gamma_1$ and $x'> \gamma_2$.
\end{enumerate}
To identify these faults accurately, classification methods can be one appropriate solution. Dempster-Shafer theory is introduced next for the detection of anomalous nodes in IoT environment.
\section{Proposed Method for Fault Detection} 
\label{pm}
The problem of fault detection in a sensor node has been considered to be a classification problem where each class is characterised by considering the different values of the data sources. For any scenario where the detection is being performed, the raw sensor data from the sensing devices are first sent to an IoT gateway where they are aggregated or fused. This fused data is then transferred to the cloud for processing the raw data and calculating the MA values which are then combined using (\ref{eq:1}) for determining the status of a node. This decision (faulty or normal) is then intimated through the Internet using a smart phone or any other relevant technology to the end-user. Based on the decision, the user will accordingly take any relevant action. 
%
\subsection{Mass Assignment Function Calculation}  
Before designing the mass assignment function, first a discussion about the sensory data behaviour becomes imperative. The proposed method is based on the assumption that the sensory data is normally distributed. Mostly used for modelling complex phenomenon in statistical methods, normal or Gaussian distribution is a very common continuous probability distribution. The reasons for assuming that the data used in this work is normally distributed are many folds. Firstly, as the number of observations in this work is sufficiently large, normal distribution is a reasonable assumption. Secondly, the normal curve has a `bell' shape which makes it ideal for modelling any system. Thirdly, when the data is normally distributed many results can be derived analytically. Moreover, normal distribution has similar properties like central limit theorem which states that under certain conditions it is possible to approximate large number of distributions to normal distribution when the sample size is sufficiently large. Now that the important properties of normal distribution have been discussed, it is necessary to ascertain if the data set used in this work actually follows normal distribution. This can be done through a \textit{probability plot}. If the data is normally distributed, then the points in the data set should lie close to a straight line indicating a normal distribution. 
\begin{figure}[H]
	\begin{subfigure}{0.5\textwidth}
	\centering
	\includegraphics[width=4.2cm,height=3.2cm]{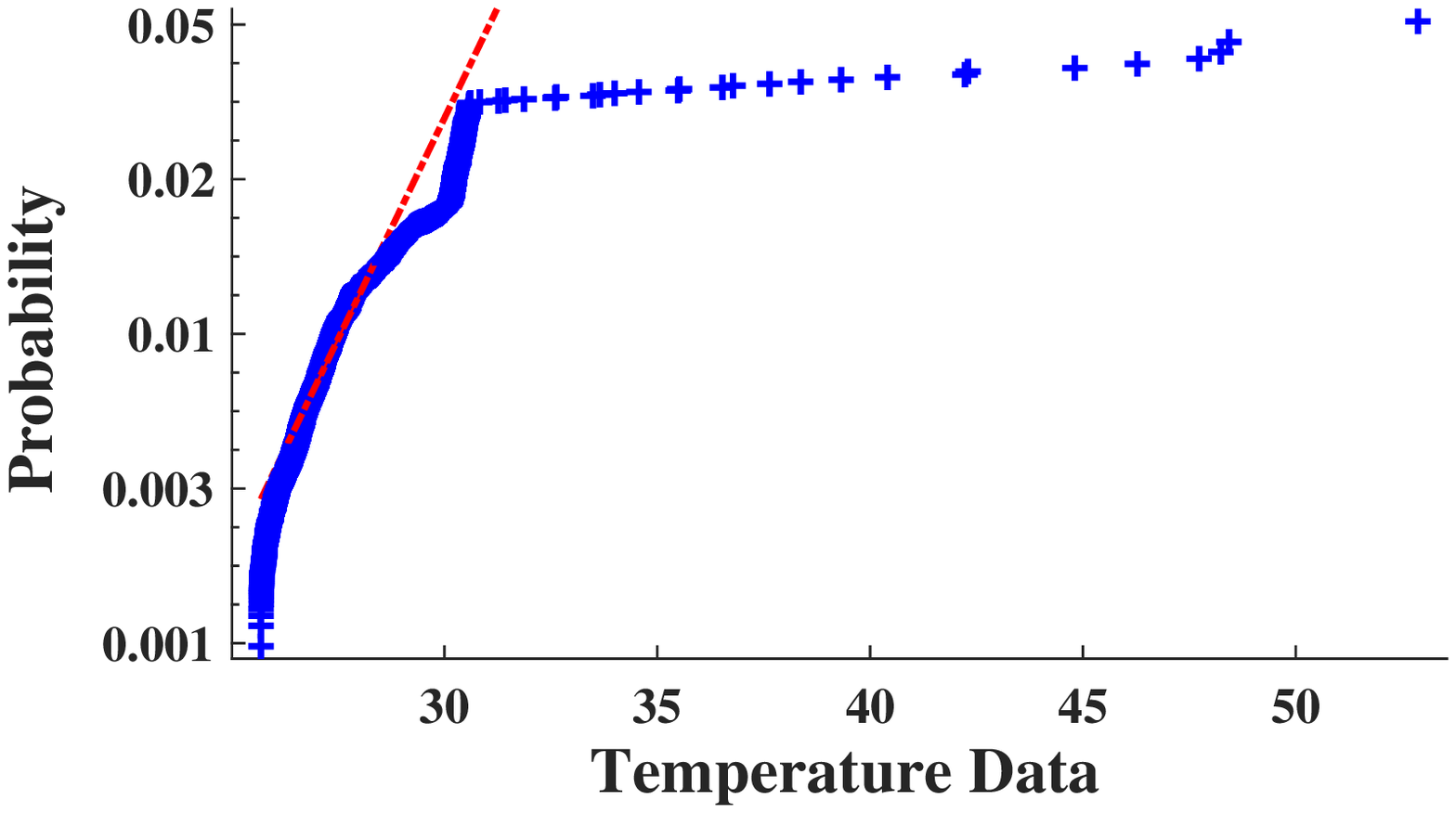}
	\label{hum}
\end{subfigure}\begin{subfigure}{0.5\textwidth}
	\centering
	\includegraphics[width=4.2cm,height=3.2cm]{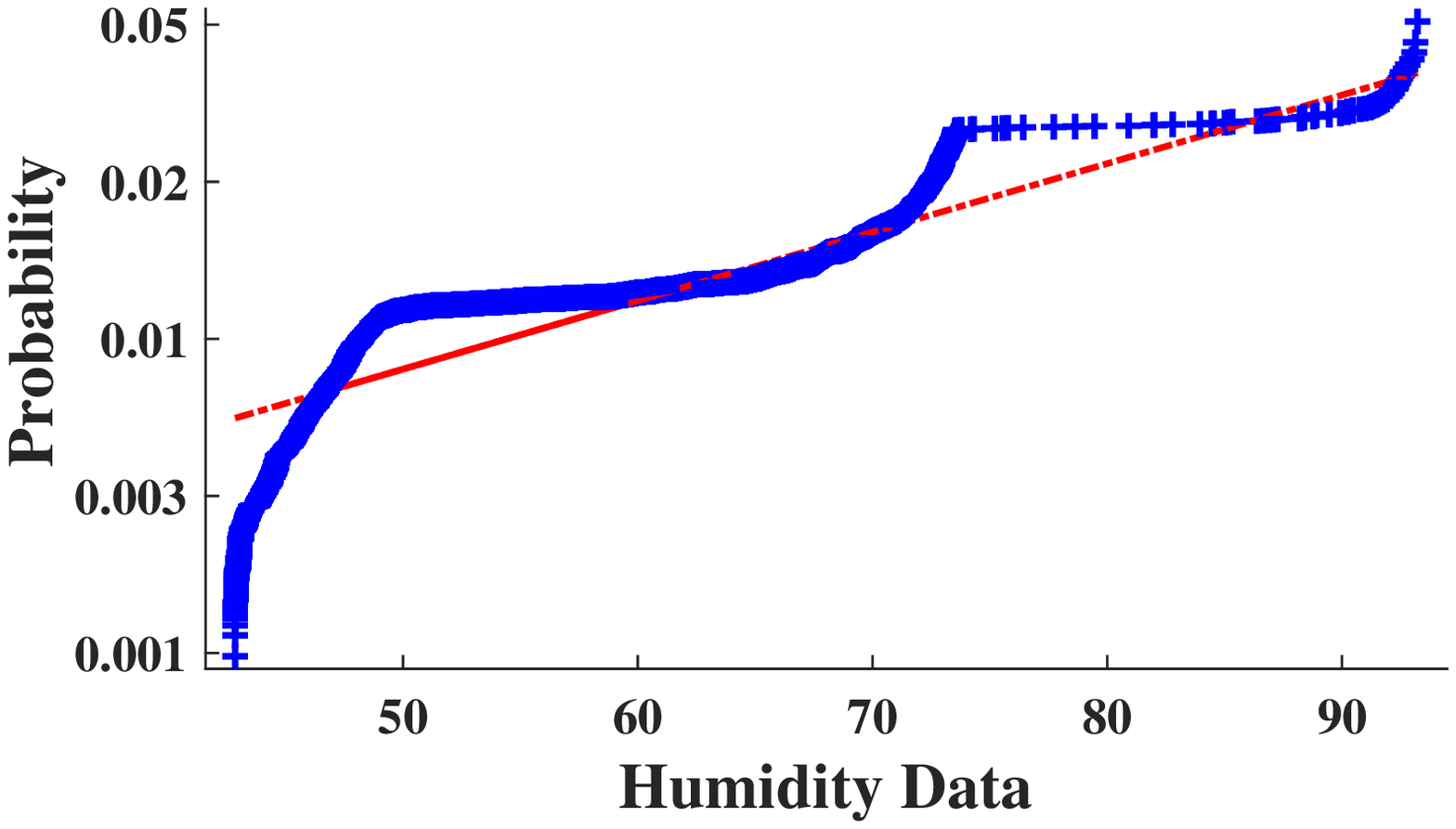}
	\label{temp}
\end{subfigure}
\caption{Probability plot for the labelled data set}\label{pp}
\end{figure}
\begin{figure}[H]
	\begin{subfigure}{0.5\textwidth}
		\centering
		\includegraphics[width=4.2cm,height=3.2cm]{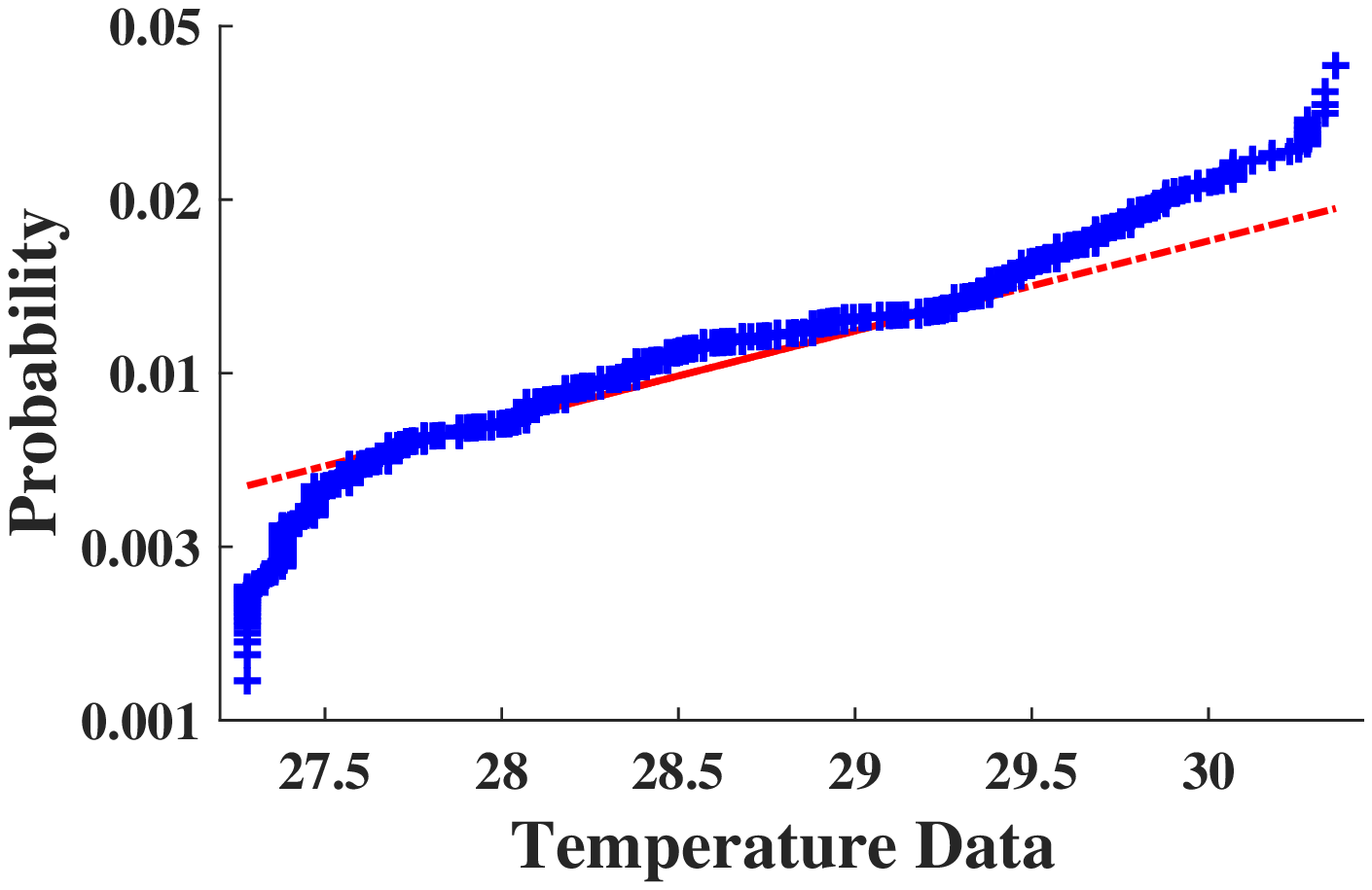}
		\label{templab}
	\end{subfigure}\begin{subfigure}{0.5\textwidth}
		\centering
		\includegraphics[width=4.2cm,height=3.2cm]{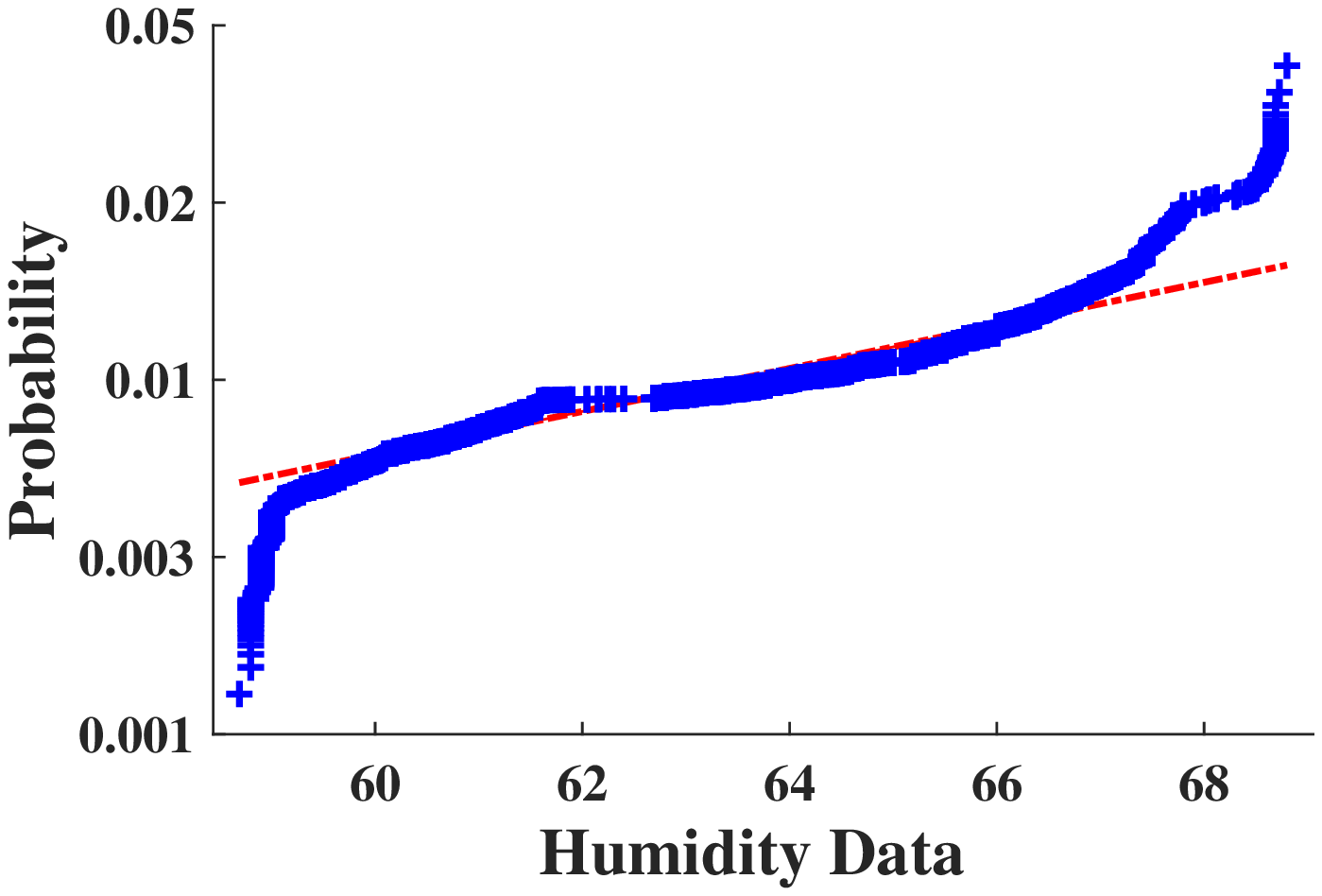}
		\label{humlab}
	\end{subfigure}
\begin{subfigure}{0.5\textwidth}
	\centering
	\includegraphics[width=4.2cm,height=3.2cm]{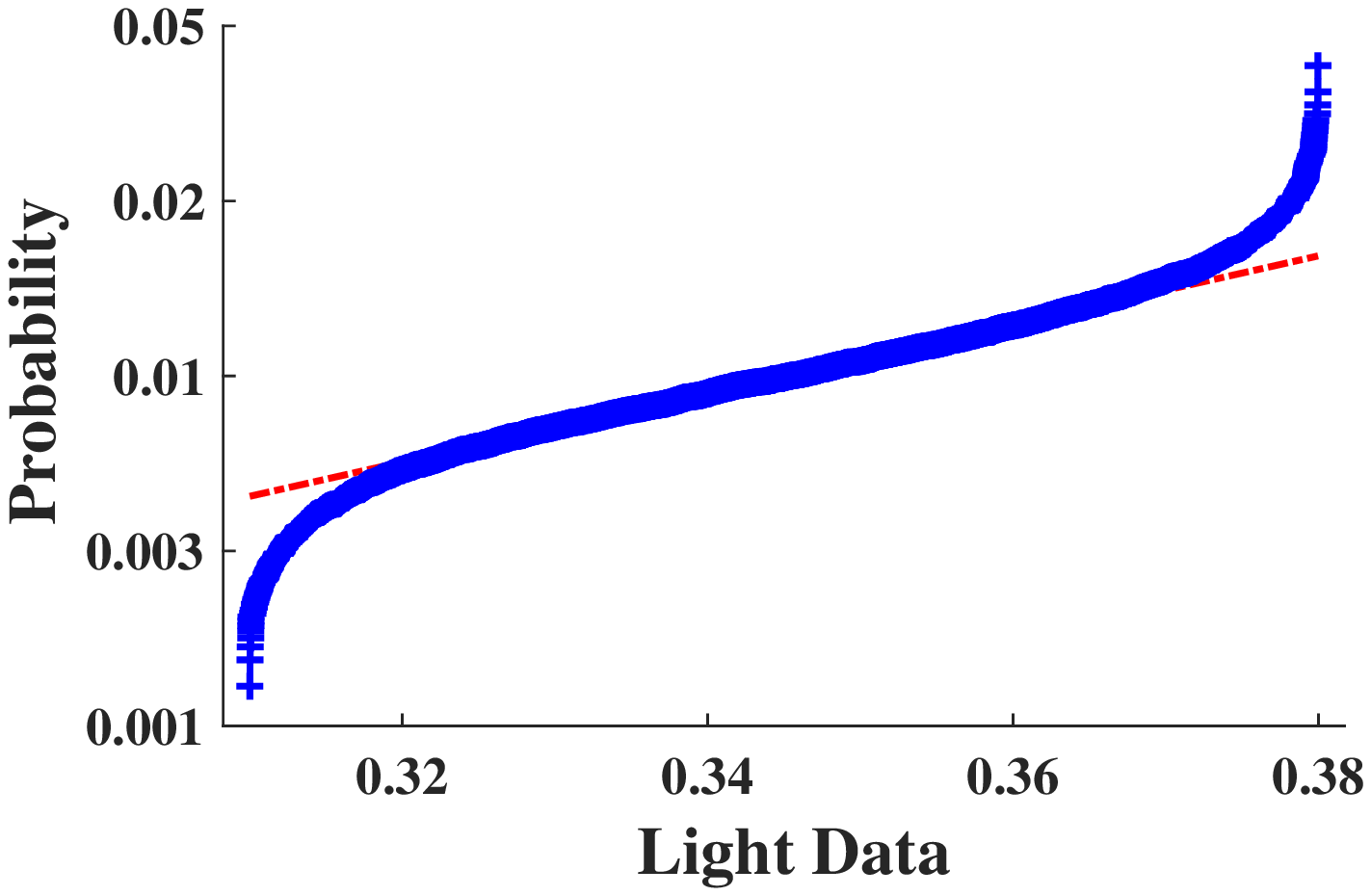}
	\label{lightlab}
\end{subfigure}
	\caption{Probability plot for the laboratory data set}\label{ld}
\end{figure}
The visual interpretation of probability plot are depicted in Fig.~\ref{pp} and Fig.~\ref{ld} from which it can be easily concluded that both the labelled and the laboratory data set follow normal distribution. The points which shows a substantive departure from the straight line are actually the outliers or the faulty data in the data set. But in general, the points lie close to the straight line thus asserting the assumption of normal distribution in this work.

Now that the data distribution is established, the mass assignment function needs to be designed. One of the most important part of Dempster-Shafer theory is the proper design of mass assignment function. In this work, the mass assignment function will depend on the underlying assumption of normal distribution. For this, at first probability density function $\rho(\delta)$ for each class $\delta = \{normal,faulty\}$ for the test vector $x_{t}$ which consists of the set of some features $\lambda$ is estimated. Probability density function is used to infer how much it is more likely that a test vector $x_{t}$ will belong to one class as compared to the other class. Based on the probability density function $\rho(\delta)$, mass assignment $m_\lambda(\delta)$ can  be given by:
\begin{equation}
m_\lambda(\delta)=\frac{\rho_\lambda(\delta)}{\sum_\delta \rho_\lambda(\delta)}
\label{ma} 
\end{equation}
where, $\rho_\lambda(\delta)$ is the probability density function and is given as:
\begin{equation}
\rho_\lambda(\delta) = \frac{1}{\sqrt{2\pi \sigma_{\delta\lambda}^2}}\exp^-{\frac{(x_{t_\lambda}-\mu_{\delta\lambda})^2}{2\sigma_{\delta\lambda}^2}}
\label{pdf} 
\end{equation}
In (\ref{pdf}), the explanation of the notations are as follows:
\begin{enumerate}
	\item [$\bullet$] $\sigma_{\delta\lambda}$ : Standard deviation of the training set for the feature $\lambda$ in class $\delta$
	\item [$\bullet$] $\mu_{\delta\lambda}$ : Expectation of the training set for the feature $\lambda$ in class $\delta$
	\item [$\bullet$] $x_{t_\lambda}$ : Value of feature $\lambda$ of the test vector $x_t$ 
\end{enumerate}
\section{Experimental Analysis}
\label{ea}
In this section, the detailed explanations of the experiments are given to show the validity of the proposed method on data collected through labelled data set~\cite{Suthaharan} and laboratory experiments. The experiments have been conducted using MATLAB R2015a. The results of the proposed method using Dempster-Shafer Theory of Evidence (DSTE) has been compared with Support Vector Machine (SVM), Classification and Regression Trees (CART) and Random Forests (RF). Different fault rates with different values of $\beta$ have been applied to the data sets.

\subsection{Steps of experiment}
To create the mass assignment function, the probability density function $\rho_\lambda(\delta)$ is first calculated for both normal and faulty data set for each test vector $x_{t}$; for every test vector $x_{t}$, $\rho(normal)$ and $\rho(faulty)$ are determined. So, for each class $\delta$ = \{normal,faulty\} and each feature $\lambda$ = \{humidity, temperature\} (and light for laboratory data), $\rho_\lambda$ for a test vector $x_t$ is calculated as:
\begin{equation}
\rho_{hum}(\delta) = \frac{1}{\sqrt{2\pi \sigma_{\delta{hum}}^2}}\exp^-{\frac{(x_{t_{hum}}-\mu_{\delta{hum}})^2}{2\sigma_{\delta{hum}}^2}}
\end{equation}
\begin{equation}
\rho_{temp}(\delta) = \frac{1}{\sqrt{2\pi \sigma_{\delta{temp}}^2}}\exp^-{\frac{(x_{t_{temp}}-\mu_{\delta{temp}})^2}{2\sigma_{\delta{temp}}^2}}
\end{equation}
\begin{equation}
\rho_{light}(\delta) = \frac{1}{\sqrt{2\pi \sigma_{\delta{light}}^2}}\exp^-{\frac{(x_{t_{light}}-\mu_{\delta{light}})^2}{2\sigma_{\delta{light}}^2}}
\end{equation}
Once these values are determined, $m_\lambda(\delta)$ is calculated using (\ref{ma}). After the calculation of  $m_\lambda(\delta)$, the mass functions are combined according to Dempster's rule of combination as given in (\ref{eq:1}). The inference regarding the state of a node is based on this combination rule. The decision is taken in favour of that class which has the higher evidence. This can be done by fusing the data obtained from the humidity and the temperature sensors (and light sensor for laboratory data set). Their mass functions are calculated for each class by applying the combination rule; $m_{hum}\oplus m_{temp}(normal)$ and $m_{hum}\oplus m_{temp}(faulty)$. If $m_{hum}\oplus m_{temp}(normal) > m_{hum}\oplus m_{temp}(faulty)$, then the node is normal, else it is faulty. Similarly, detection of normal and faulty node is also carried out by taking the light sensor value for laboratory data set.

\subsection{Description of data sets}

\subsubsection{Description of the labelled data set}
DSTE has also been applied on a popular labelled data set to verify its applicability. The labelled data set was prepared by the researchers of University of North Carolina at Greensboro~\cite{Suthaharan}. The data consists of humidity and temperature measurements collected with the help of TelosB motes measured every 5 seconds for 6 hours. The data set was prepared with single-hop and multi-hop networks. Anomalies were introduced in the data set with the help of water kettle to change the humidity and temperature values; thus creating two classes for the data sets, normal and faulty. To apply the proposed method and the compared methods, the four types of faults as described in Section~\ref{fault} have been incorporated in the data set for multi-hop networks; 1 for anomalous data and 0 for normal data. A total of 18760 observations has been aggregated to carry out the experiments. Different percentage of faulty nodes (fault rates) have been considered to prepare the final data set. 

\subsubsection{Data collected through laboratory setup}
For further verification, experimental analysis through laboratory setup have been conducted where data is recorded using temperature, humidity and light sensors by using an arduino microcontroller system. The data were recorded in the university premises in the month of June 2019, from 9:15 in the morning till 17:30 in the evening at an interval of 13 seconds. A total of 2556 observations were recorded with three parameters. As mentioned before, faults were introduced in the data set. For all the observations, temperature, humidity and light values along with the classification category (1 for faulty and 0 for normal) and the time stamps are described.
\subsection{Results}
Validation Set has approach been used in this work to validate the proposed method. In validation set approach, the data set is divided into two parts; one part is called the \textit{validation set} or popularly known as the \textit{hold-out set} or simply the \textit{test set} and the other part is the \textit{training set}. Out of the total observations, 70\% of the data is considered as the training set and the rest is used as a test data set. 

Quantitative metrics have been used in this work for the exhaustive evaluation of the Dempster-Shafer based classifier. For classification problems, let the ground truth present in the data set $\Delta$ be $\zeta$ and the inferred faulty nodes be $\eta$. Let at any time instance $t$, the inferred faulty nodes be $\eta_0(t)$ and the normal nodes be $\eta_1(t)$. From this knowledge, the following definitions can be deduced~\cite{Aggarwal2015}.
 \begin{itemize}
	\item Accuracy: Accuracy gives the correctly identified normal and faulty nodes in the total data set. Accuracy can be given as:
	\begin{equation}
	Accuracy = \frac{\mid(\eta_0(t)\cap \zeta)\mid\cup \mid(\eta_1(t)\cap \zeta)\mid}{\mid\Delta \mid}
	\label{eq:6}
	\end{equation} 
	\item Sensitivity or Recall (True Positive Rate): Sensitivity or Recall is the percentage of correctly predicted faulty nodes from the total reported faulty nodes in the data set. 
	\begin{equation}
	Sensitivity= \frac{\mid\eta_0(t)\cap \zeta\mid}{\mid\zeta\mid}
	\label{eq:7}
	\end{equation}
		\item Specificity (True Negative Rate): Specificity is the percentage of correctly identified normal nodes from the set of normal nodes in the data set.
	\begin{equation}
	Specificity = \frac{\mid\eta_1(t)\cap \zeta\mid}{\mid\zeta\mid}
	\label{eq:8}
	\end{equation}
	\item False Positive Rate (FPR): FPR is the percentage of the falsely reported positives from the negatives in $\zeta$.
	 \begin{equation}
	   FPR = \frac{\mid\eta_0(t)-\zeta\mid}{\mid\Delta-\zeta\mid}
	   \label{eq:9}
	  \end{equation}
	\item Precision: Precision is the percentage of correctly predicted faulty nodes from the total set of reported faulty nodes.
	  \begin{equation}
	   Precision = \frac{\mid\eta_0(t)\cap \zeta \mid}{\mid\eta_0(t)\mid}
	   \label{eq:10}
	    \end{equation}

\end{itemize}
The range of each of the parameters lies between [0,1]; for Accuracy, Sensitivity, Specificity and Precision, a higher value indicates better results. On the contrary, for FPR, a lower value is desired.

\subsubsection{Analysis of labelled data set}
The results and analysis based on the different categories of faults are described next. Table~\ref{tab_1} gives the values for the different metrics based on different kinds of faults for DSTE, SVM, CART and RF which consider the information from both the temperature and the humidity sensor. 

	\begin{table*}
	\setlength{\tabcolsep}{5.5pt}
	\vspace{-0.9em}    
	\caption{Results for Faulty Node Analysis. The best values are given in bold.}
	\label{tab_1}
	\resizebox{0.7\textwidth}{!}{
		\begin{tabular}{l|l|l|l|l|l|l|l}
			\cline{2-7}

			& \multicolumn{1}{|c|} {Fault Rate} & \multicolumn{5}{|c|} {10\%} &\\
			\cline{2-7}
			& $\beta$ & 2 & 4 & 6 & 8 & 10 & \\
			\cline{1-8}
			\multicolumn{1}{|c|}{}  & \multicolumn{1}{|c|} {}& \multicolumn{5}{|c|} {79.56} & \multicolumn{1}{|c|}{DSTE}\\
			\cline{3-8}
			\multicolumn{1}{|c|}{} & \multicolumn{1}{|c|}{Offset Fault} & \multicolumn{5}{|c|} {\textbf{79.71}} & \multicolumn{1}{|c|}{SVM}\\
			\cline{3-8}
			\multicolumn{1}{|c|}{} & \multicolumn{1}{|c|}{} & \multicolumn{5}{|c|} {74.2} & \multicolumn{1}{|c|}{CART}\\
			\cline{3-8}
			\multicolumn{1}{|c|}{} & \multicolumn{1}{|c|}{}  & \multicolumn{5}{|c|} {75.76} & \multicolumn{1}{|c|}{RF}\\
			\cline{2-8}
			\multicolumn{1}{|c|}{Accuracy(\%)} &  & \textbf{99.56} & \textbf{99.27} & \textbf{99.23} & \textbf{99.23} & \textbf{99.23} & \multicolumn{1}{|c|}{DSTE}\\
			\cline{3-8}
			\multicolumn{1}{|c|}{} & \multicolumn{1}{|c|}{Gain Fault} & 79.40 & 78.99 & 78.83 & 78.74 & 78.71 & \multicolumn{1}{|c|}{SVM}\\
			\cline{3-8}
			\multicolumn{1}{|c|}{} & \multicolumn{1}{|c|}{} & 97.67 & 97.67 & 97.67 & 97.67 & 97.67 & \multicolumn{1}{|c|}{CART}\\
			\cline{3-8}
			\multicolumn{1}{|c|}{} & \multicolumn{1}{|c|}{} & 99.23 & 98.86 & 98.86 & 98.95 & 99.11 & \multicolumn{1}{|c|}{RF}\\
			\cline{2-8}
			\multicolumn{1}{|c|}{}  & \multicolumn{1}{|c|} {}& \multicolumn{5}{|c|} {\textbf{99.23}} & \multicolumn{1}{|c|}{DSTE}\\
			\cline{3-8}
			\multicolumn{1}{|c|}{} & \multicolumn{1}{|c|}{Out-of-Bounds Fault} & \multicolumn{5}{|c|} {79.71} & \multicolumn{1}{|c|}{SVM}\\
			\cline{3-8}
			\multicolumn{1}{|c|}{} & \multicolumn{1}{|c|}{} & \multicolumn{5}{|c|} {97.67} & \multicolumn{1}{|c|}{CART}\\
			\cline{3-8}
			\multicolumn{1}{|c|}{} & \multicolumn{1}{|c|}{}  & \multicolumn{5}{|c|} {99.11} & \multicolumn{1}{|c|}{RF}\\
			\cline{2-8}
			\multicolumn{1}{|c|}{}  & \multicolumn{1}{|c|} {}& \multicolumn{5}{|c|} {\textbf{99.23}} & \multicolumn{1}{|c|}{DSTE}\\
			\cline{3-8}
			\multicolumn{1}{|c|}{} & \multicolumn{1}{|c|}{Data-Loss Fault} & \multicolumn{5}{|c|} {81.66} & \multicolumn{1}{|c|}{SVM}\\
			\cline{3-8}
			\multicolumn{1}{|c|}{} & \multicolumn{1}{|c|}{} & \multicolumn{5}{|c|} {97.67} & \multicolumn{1}{|c|}{CART}\\
			\cline{3-8}
			\multicolumn{1}{|c|}{} & \multicolumn{1}{|c|}{}  & \multicolumn{5}{|c|} {99.11} & \multicolumn{1}{|c|}{RF}\\
			\cline{1-8}
			\multicolumn{1}{|c|}{}  & \multicolumn{1}{|c|} {}& \multicolumn{5}{|c|} {\textbf{22.29}} & \multicolumn{1}{|c|}{DSTE}\\
			\cline{3-8}
			\multicolumn{1}{|c|}{} & \multicolumn{1}{|c|}{Offset Fault} & \multicolumn{5}{|c|} {22.55} & \multicolumn{1}{|c|}{SVM}\\
			\cline{3-8}
			\multicolumn{1}{|c|}{} & \multicolumn{1}{|c|}{} & \multicolumn{5}{|c|} {28.76} & \multicolumn{1}{|c|}{CART}\\
			\cline{3-8}
			\multicolumn{1}{|c|}{} & \multicolumn{1}{|c|}{}  & \multicolumn{5}{|c|} {26.99} & \multicolumn{1}{|c|}{RF}\\
			\cline{2-8}
			\multicolumn{1}{|c|}{FPR(\%)} &  & \textbf{0} & \textbf{0} & \textbf{0} & \textbf{0} & \textbf{0} & \multicolumn{1}{|c|}{DSTE}\\
			\cline{3-8}
			\multicolumn{1}{|c|}{} & \multicolumn{1}{|c|}{Gain Fault} & 22.89 & 23.35 & 23.53 & 23.63 & 23.67 & \multicolumn{1}{|c|}{SVM}\\
			\cline{3-8}
			\multicolumn{1}{|c|}{} & \multicolumn{1}{|c|}{} & 2.28 & 2.28 & 2.28 & 2.28 & 2.28 & \multicolumn{1}{|c|}{CART}\\
			\cline{3-8}
			\multicolumn{1}{|c|}{} & \multicolumn{1}{|c|}{} & 0.52 & 0.94 & 0.94 & 0.84 & 0.66 & \multicolumn{1}{|c|}{RF}\\
			\cline{2-8}
			\multicolumn{1}{|c|}{}  & \multicolumn{1}{|c|} {}& \multicolumn{5}{|c|} {\textbf{0}} & \multicolumn{1}{|c|}{DSTE}\\
			\cline{3-8}
			\multicolumn{1}{|c|}{} & \multicolumn{1}{|c|}{Out-of-Bounds Fault} & \multicolumn{5}{|c|} {22.55} & \multicolumn{1}{|c|}{SVM}\\
			\cline{3-8}
			\multicolumn{1}{|c|}{} & \multicolumn{1}{|c|}{} & \multicolumn{5}{|c|} {2.28} & \multicolumn{1}{|c|}{CART}\\
			\cline{3-8}
			\multicolumn{1}{|c|}{} & \multicolumn{1}{|c|}{}  & \multicolumn{5}{|c|} {0.05} & \multicolumn{1}{|c|}{RF}\\
			\cline{2-8}
			\multicolumn{1}{|c|}{}  & \multicolumn{1}{|c|} {}& \multicolumn{5}{|c|} {\textbf{0}} & \multicolumn{1}{|c|}{DSTE}\\
			\cline{3-8}
			\multicolumn{1}{|c|}{} & \multicolumn{1}{|c|}{Data-Loss Fault} & \multicolumn{5}{|c|} {20.34} & \multicolumn{1}{|c|}{SVM}\\
			\cline{3-8}
			\multicolumn{1}{|c|}{} & \multicolumn{1}{|c|}{} & \multicolumn{5}{|c|} {2.28} & \multicolumn{1}{|c|}{CART}\\
			\cline{3-8}
			\multicolumn{1}{|c|}{} & \multicolumn{1}{|c|}{}  & \multicolumn{5}{|c|} {0.76} & \multicolumn{1}{|c|}{RF}\\
			\cline{1-8}
			\multicolumn{1}{|c|}{}  & \multicolumn{1}{|c|} {}& \multicolumn{5}{|c|} {35.08} & \multicolumn{1}{|c|}{DSTE}\\
			\cline{3-8}
			\multicolumn{1}{|c|}{} & \multicolumn{1}{|c|}{Offset Fault} & \multicolumn{5}{|c|} {\textbf{35.6}} & \multicolumn{1}{|c|}{SVM}\\
			\cline{3-8}
			\multicolumn{1}{|c|}{} & \multicolumn{1}{|c|}{} & \multicolumn{5}{|c|} {30.23} & \multicolumn{1}{|c|}{CART}\\
			\cline{3-8}
			\multicolumn{1}{|c|}{} & \multicolumn{1}{|c|}{}  & \multicolumn{5}{|c|} {31.59} & \multicolumn{1}{|c|}{RF}\\
			\cline{2-8}
			\multicolumn{1}{|c|}{Precision(\%)} &  &\textbf{100} & \textbf{100} & \textbf{100} & \textbf{100} & \textbf{100} & \multicolumn{1}{|c|}{DSTE}\\
			\cline{3-8}
			\multicolumn{1}{|c|}{} & \multicolumn{1}{|c|}{Gain Fault} & 35.26 & 34.8 & 34.63 & 34.53 & 34.5 & \multicolumn{1}{|c|}{SVM}\\
			\cline{3-8}
			\multicolumn{1}{|c|}{} & \multicolumn{1}{|c|}{} & 84.51 & 84.51 & 84.51 & 84.51 & 84.51 & \multicolumn{1}{|c|}{CART}\\
			\cline{3-8}
			\multicolumn{1}{|c|}{} & \multicolumn{1}{|c|}{} & 95.98 & 92.97 & 92.97 & 93.67 & 94.97 & \multicolumn{1}{|c|}{RF}\\
			\cline{2-8}
			\multicolumn{1}{|c|}{}  & \multicolumn{1}{|c|} {}& \multicolumn{5}{|c|} {\textbf{100}} & \multicolumn{1}{|c|}{DSTE}\\
			\cline{3-8}
			\multicolumn{1}{|c|}{} & \multicolumn{1}{|c|}{Out-of-Bounds Fault} & \multicolumn{5}{|c|} {35.6} & \multicolumn{1}{|c|}{SVM}\\
			\cline{3-8}
			\multicolumn{1}{|c|}{} & \multicolumn{1}{|c|}{} & \multicolumn{5}{|c|} {84.51} & \multicolumn{1}{|c|}{CART}\\
			\cline{3-8}
			\multicolumn{1}{|c|}{} & \multicolumn{1}{|c|}{}  & \multicolumn{5}{|c|} {94.96} & \multicolumn{1}{|c|}{RF}\\
			\cline{2-8}
			\multicolumn{1}{|c|}{}  & \multicolumn{1}{|c|} {}& \multicolumn{5}{|c|} {\textbf{100}} & \multicolumn{1}{|c|}{DSTE}\\
			\cline{3-8}
			\multicolumn{1}{|c|}{} & \multicolumn{1}{|c|}{Data-Loss Fault} & \multicolumn{5}{|c|} {37.99} & \multicolumn{1}{|c|}{SVM}\\
			\cline{3-8}
			\multicolumn{1}{|c|}{} & \multicolumn{1}{|c|}{} & \multicolumn{5}{|c|} {84.51} & \multicolumn{1}{|c|}{CART}\\
			\cline{3-8}
			\multicolumn{1}{|c|}{} & \multicolumn{1}{|c|}{}  & \multicolumn{5}{|c|} {95.98} & \multicolumn{1}{|c|}{RF}\\
			\cline{1-8}
			\multicolumn{1}{|c|}{}  & \multicolumn{1}{|c|} {}& \multicolumn{5}{|c|} {94.05} & \multicolumn{1}{|c|}{DSTE}\\
			\cline{3-8}
			\multicolumn{1}{|c|}{} & \multicolumn{1}{|c|}{Offset Fault} & \multicolumn{5}{|c|} {\textbf{97.34}} & \multicolumn{1}{|c|}{SVM}\\
			\cline{3-8}
			\multicolumn{1}{|c|}{} & \multicolumn{1}{|c|}{} & \multicolumn{5}{|c|} {\textbf{97.34}} & \multicolumn{1}{|c|}{CART}\\
			\cline{3-8}
			\multicolumn{1}{|c|}{} & \multicolumn{1}{|c|}{}  & \multicolumn{5}{|c|} {\textbf{97.34}} & \multicolumn{1}{|c|}{RF}\\
			\cline{2-8}
			\multicolumn{1}{|c|}{Sensitivity(\%)} &  &96.08 & 93.58 & 93.27 & 93.27 & 93.27 & \multicolumn{1}{|c|}{DSTE}\\
			\cline{3-8}
			\multicolumn{1}{|c|}{} & \multicolumn{1}{|c|}{Gain Fault} & \textbf{97.34} & \textbf{97.34} & \textbf{97.34} & \textbf{97.34} & \textbf{97.34} & \multicolumn{1}{|c|}{SVM}\\
			\cline{3-8}
			\multicolumn{1}{|c|}{} & \multicolumn{1}{|c|}{} & \textbf{97.34} & \textbf{97.34} & \textbf{97.34} & \textbf{97.34} & \textbf{97.34} & \multicolumn{1}{|c|}{CART}\\
			\cline{3-8}
			\multicolumn{1}{|c|}{} & \multicolumn{1}{|c|}{} & \textbf{97.34} & \textbf{97.34} & \textbf{97.34} & \textbf{97.34} & \textbf{97.34} & \multicolumn{1}{|c|}{RF}\\
			\cline{2-8}
			\multicolumn{1}{|c|}{}  & \multicolumn{1}{|c|} {}& \multicolumn{5}{|c|} {93.27} & \multicolumn{1}{|c|}{DSTE}\\
			\cline{3-8}
			\multicolumn{1}{|c|}{} & \multicolumn{1}{|c|}{Out-of-Bounds Fault} & \multicolumn{5}{|c|} {\textbf{97.34}} & \multicolumn{1}{|c|}{SVM}\\
			\cline{3-8}
			\multicolumn{1}{|c|}{} & \multicolumn{1}{|c|}{} & \multicolumn{5}{|c|} {\textbf{97.34}} & \multicolumn{1}{|c|}{CART}\\
			\cline{3-8}
			\multicolumn{1}{|c|}{} & \multicolumn{1}{|c|}{}  & \multicolumn{5}{|c|} {\textbf{97.34}} & \multicolumn{1}{|c|}{RF}\\
			\cline{2-8}
			\multicolumn{1}{|c|}{}  & \multicolumn{1}{|c|} {}& \multicolumn{5}{|c|} {93.27} & \multicolumn{1}{|c|}{DSTE}\\
			\cline{3-8}
			\multicolumn{1}{|c|}{} & \multicolumn{1}{|c|}{Data-Loss Fault} & \multicolumn{5}{|c|} {\textbf{97.34}} & \multicolumn{1}{|c|}{SVM}\\
			\cline{3-8}
			\multicolumn{1}{|c|}{} & \multicolumn{1}{|c|}{} & \multicolumn{5}{|c|} {\textbf{97.34}} & \multicolumn{1}{|c|}{CART}\\
			\cline{3-8}
			\multicolumn{1}{|c|}{} & \multicolumn{1}{|c|}{}  & \multicolumn{5}{|c|} {\textbf{97.34}} & \multicolumn{1}{|c|}{RF}\\
			\cline{1-8}
			\multicolumn{1}{|c|}{}  & \multicolumn{1}{|c|} {}& \multicolumn{5}{|c|} {\textbf{77.71}} & \multicolumn{1}{|c|}{DSTE}\\
			\cline{3-8}
			\multicolumn{1}{|c|}{} & \multicolumn{1}{|c|}{Offset Fault} & \multicolumn{5}{|c|} {77.45} & \multicolumn{1}{|c|}{SVM}\\
			\cline{3-8}
			\multicolumn{1}{|c|}{} & \multicolumn{1}{|c|}{} & \multicolumn{5}{|c|} {71.23} & \multicolumn{1}{|c|}{CART}\\
			\cline{3-8}
			\multicolumn{1}{|c|}{} & \multicolumn{1}{|c|}{}  & \multicolumn{5}{|c|} {72.86} & \multicolumn{1}{|c|}{RF}\\
			\cline{2-8}
			\multicolumn{1}{|c|}{Specificity(\%)} &  &\textbf{100}& \textbf{100} & \textbf{100} & \textbf{100} & \textbf{100} & \multicolumn{1}{|c|}{DSTE}\\
			\cline{3-8}
			\multicolumn{1}{|c|}{} & \multicolumn{1}{|c|}{Gain Fault} &  77.10 & 76.64 & 76.46 & 76.36 & 76.32 & \multicolumn{1}{|c|}{SVM}\\
			\cline{3-8}
			\multicolumn{1}{|c|}{} & \multicolumn{1}{|c|}{} & 97.71 & 97.71 & 97.71 & 97.71 & 97.71 & \multicolumn{1}{|c|}{CART}\\
			\cline{3-8}
			\multicolumn{1}{|c|}{} & \multicolumn{1}{|c|}{} & 99.33 & 99.48 & 99.05 & 99.05 & 99.16 & \multicolumn{1}{|c|}{RF}\\
			\cline{2-8}
			\multicolumn{1}{|c|}{}  & \multicolumn{1}{|c|} {}& \multicolumn{5}{|c|} {\textbf{100}} & \multicolumn{1}{|c|}{DSTE}\\
			\cline{3-8}
			\multicolumn{1}{|c|}{} & \multicolumn{1}{|c|}{Out-of-Bounds Fault} & \multicolumn{5}{|c|} {77.45} & \multicolumn{1}{|c|}{SVM}\\
			\cline{3-8}
			\multicolumn{1}{|c|}{} & \multicolumn{1}{|c|}{} & \multicolumn{5}{|c|} {97.71} & \multicolumn{1}{|c|}{CART}\\
			\cline{3-8}
			\multicolumn{1}{|c|}{} & \multicolumn{1}{|c|}{}  & \multicolumn{5}{|c|} {99.19} & \multicolumn{1}{|c|}{RF}\\
			\cline{2-8}
			\multicolumn{1}{|c|}{}  & \multicolumn{1}{|c|} {}& \multicolumn{5}{|c|} {\textbf{100}} & \multicolumn{1}{|c|}{DSTE}\\
			\cline{3-8}
			\multicolumn{1}{|c|}{} & \multicolumn{1}{|c|}{Data-Loss Fault} & \multicolumn{5}{|c|} {79.65} & \multicolumn{1}{|c|}{SVM}\\
			\cline{3-8}
			\multicolumn{1}{|c|}{} & \multicolumn{1}{|c|}{} & \multicolumn{5}{|c|} {97.71} & \multicolumn{1}{|c|}{CART}\\
			\cline{3-8}
			\multicolumn{1}{|c|}{} & \multicolumn{1}{|c|}{}  & \multicolumn{5}{|c|} {99.33} & \multicolumn{1}{|c|}{RF}\\
			\cline{1-8}
		\end{tabular}
	}
\end{table*}

It can be seen from Table~\ref{tab_1} that for the different types of faults, DSTE gives the best accuracy in most of the cases, other than the offset fault which is also quite close to the best accuracy as exhibited by SVM. DSTE shows very competitive results for false positive rate wherein for gain, out-of-bounds and data faults, it has an FPR of 0 which somewhat confirms the validity of the proposed method. As compared to the other methods, the precision rate is also quite high for DSTE. However, DSTE has a comparatively lower value of sensitivity but a very high value of specificity which may confuse the readers as to the authenticity of the proposed method. Thus, there are mixed results when all the metrics are considered together. To overcome this confusion and to determine which statistical method has the best classification ability Area under an Receiver Operating
Characteristic (ROC) Curve has been taken into consideration. Area under an Receiver Operating Characteristic (AUROC) (also known as AUC (Area Under the Curve) ROC (Receiver Operating Characteristics)) exhibits the performance measurement of classification problems at different threshold settings. The curve is plotted with True Positive Rate (Sensitivity) against the False Positive Rate. It has the capability to determine the class distinguishing ability of a classifier. A higher value of the area under the curve for a classifier is an indication of its superior performance. With a value near to 1, a classifier represents a good measure of separability.  

 \begin{figure}
	\begin{subfigure}{0.5\textwidth}
		\centering
		\includegraphics[width=4cm,height=3.5cm]{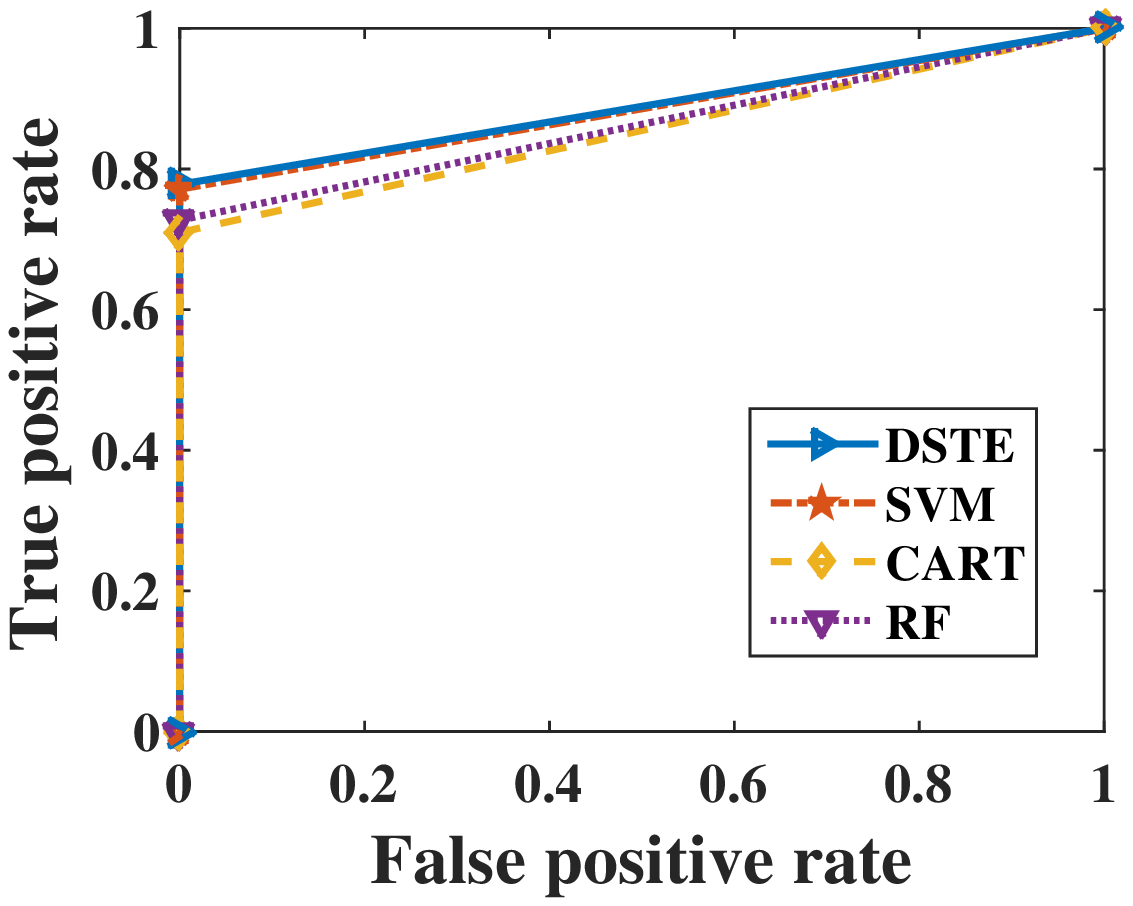}
		\caption{ROC for offset fault}
		\label{off}
	\end{subfigure}\begin{subfigure}{0.5\textwidth}
		\centering
		\includegraphics[width=4cm,height=3.5cm]{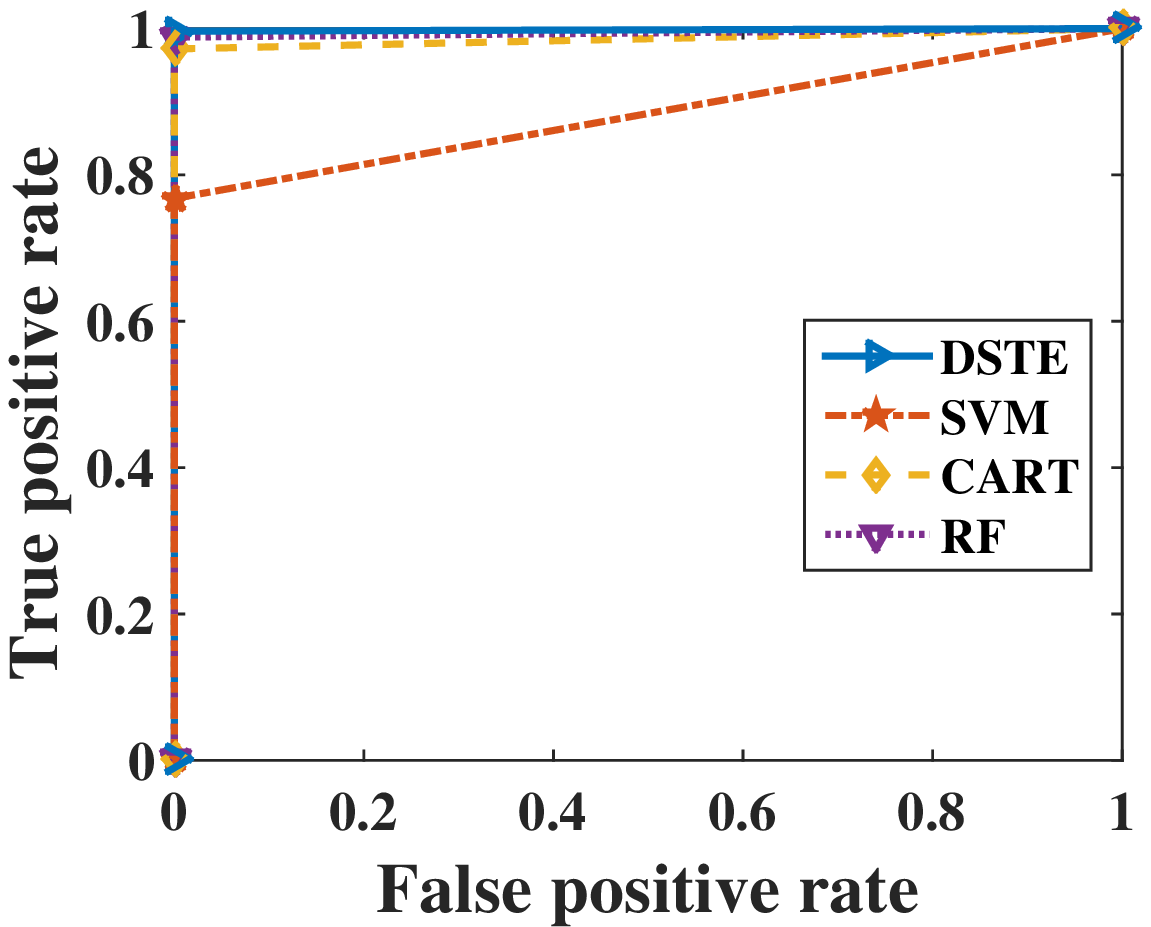}
		\caption{ROC for gain fault}
		\label{gain}
	\end{subfigure}
	\begin{subfigure}{0.5\textwidth}
		\centering
		\includegraphics[width=4cm,height=3.5cm]{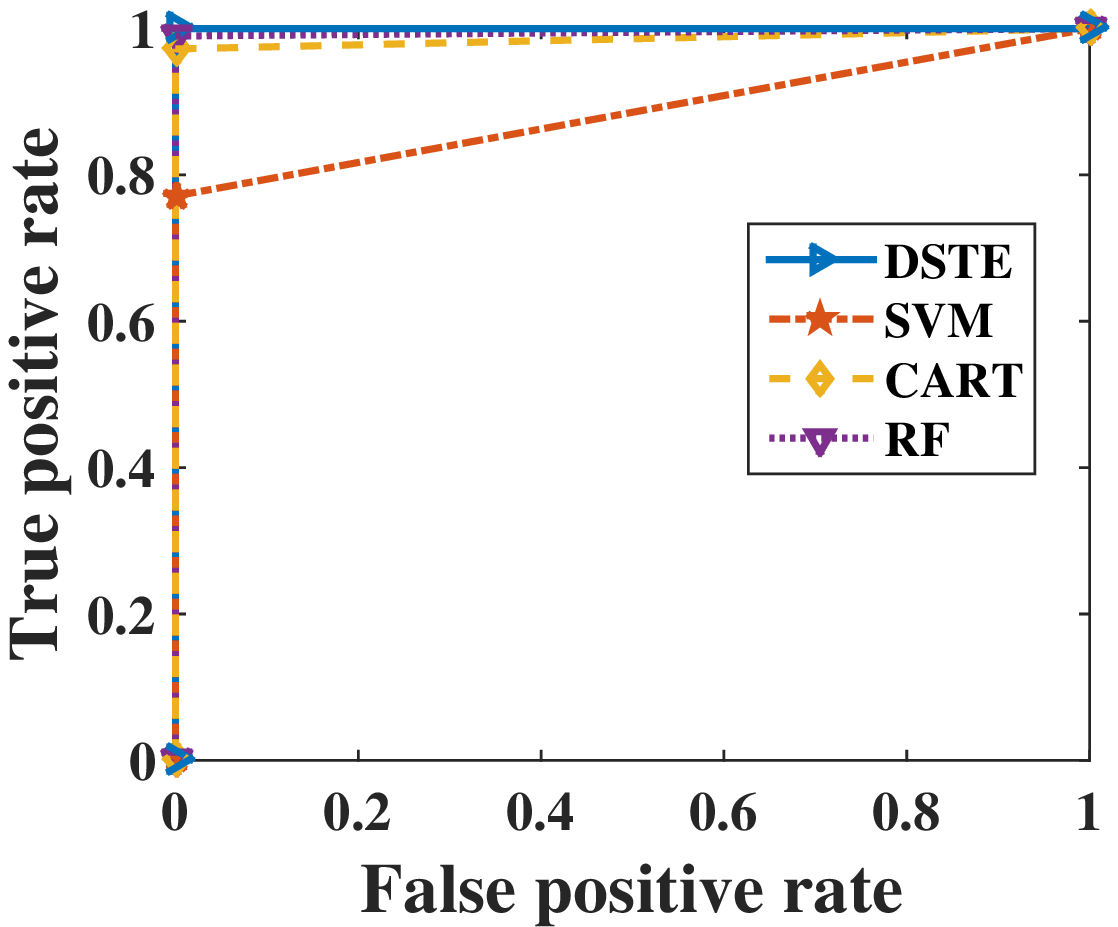}
		\caption{ROC for out-of-bounds fault}
		\label{oob}
	\end{subfigure}\begin{subfigure}{0.5\textwidth}
		\centering
		\includegraphics[width=4cm,height=3.5cm]{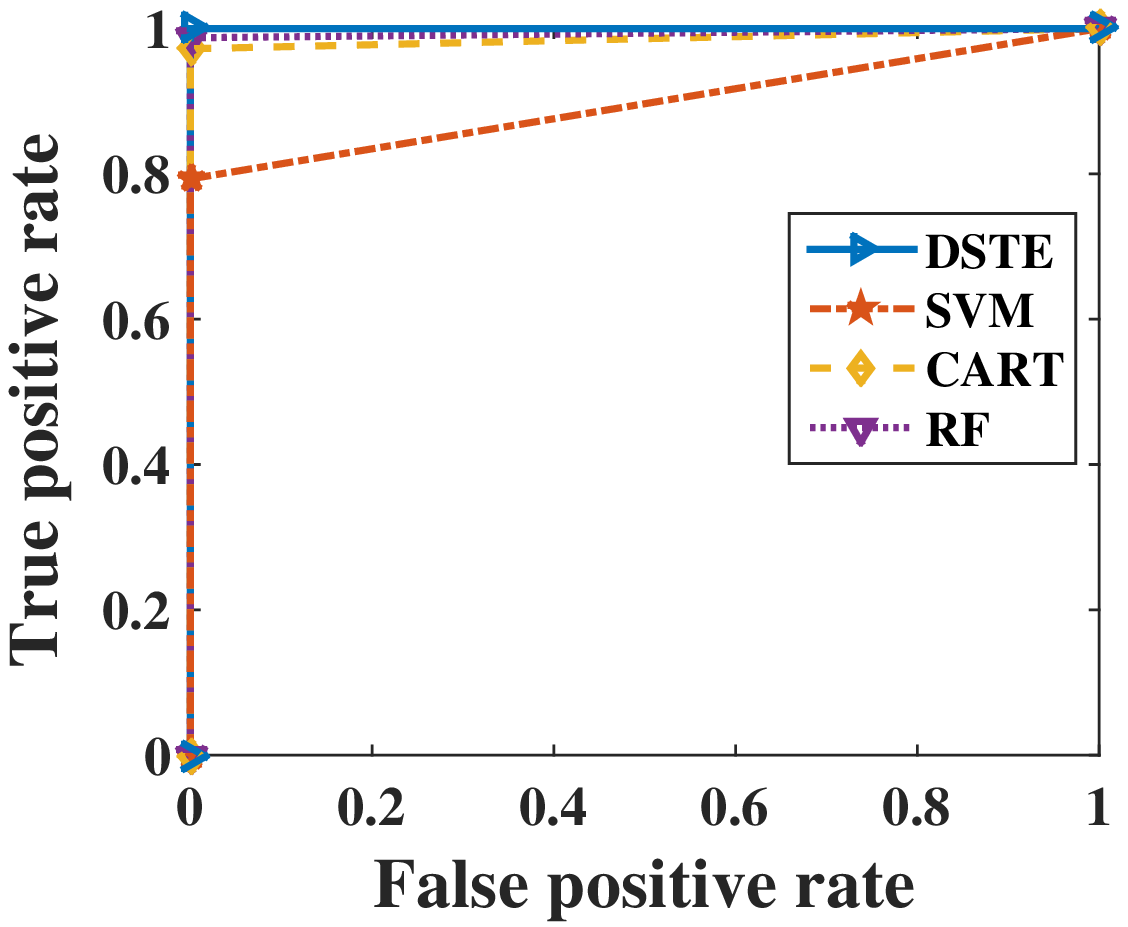}
		\caption{ROC for data-loss fault}
		\label{dl}
	\end{subfigure}
	\caption{ROC for different types of faults (Fault rate =10\%, $\alpha=10$ and $\beta=2$)}\label{roc}
\end{figure}

\begin{table}
	\setlength{\tabcolsep}{5.5pt}
	\vspace{-0.9em}    
	\caption{Area under the Curve (AUC) for the Labelled Dataset}
	\label{tab_3}
	\resizebox{1\textwidth}{!}{
		\begin{tabular}{l l l l l}
			\cline{1-5}

			Learning Method & Offset Fault & Gain Fault & Out-of-Bounds Fault & Data-Loss Fault\\
			\cline{1-5}
			DSTE & \textbf{0.889} & \textbf{0.9982} & \textbf{1} & \textbf{1}\\
			SVM & 0.8556 & 0.8839 & 0.9861 & 0.9941\\
			CART & 0.8548 & 0.9861 & 0.9861 & 0.9861\\
			RF & 0.8636 & 0.9941 & 0.9949 & 0.9936\\
			\cline{1-5}
		\end{tabular}
	}
\end{table}
Fig.~\ref{roc} shows the ROC curve for the different faults for the proposed and the compared methods with Table~\ref{tab_3} giving the AUC values. Going by the working knowledge of AUCROC, the figures are a clear indication that DSTE outperforms all the other state-of-the-art learning methods in terms of detection of faulty nodes.  

Table~\ref{tab_2} gives the values for comparison for accuracy between the state-of-the-art statistical learning techniques with different rates of incorporated faults in the network. It can be concluded from the table that the proposed method shows the best accuracy among all the compared statistical learning techniques.

 	\begin{table}
 	\setlength{\tabcolsep}{5.5pt}
 	\vspace{-0.9em}    
 	\caption{Comparison of Accuracy for the Different Statistical Methods. The best values are given in bold.}
 	\label{tab_2}\tiny
 	\resizebox{1\textwidth}{!}{
 		\begin{tabular}{|l|l|l|l|l|l|l}
 			\cline{1-6}

 			 \multicolumn{1}{|c|} {Fault Rate} & \multicolumn{5}{|c|} {\textbf{20\%}} &\\
 			\cline{1-6}
 			 $\beta$ & 2 & 4 & 6 & 8 & 10\\
 			\cline{1-7}
 			 \multicolumn{1}{|c|} {}& \multicolumn{5}{|c|} {\textbf{90.47}} & \multicolumn{1}{|c|}{DSTE}\\
 			\cline{2-7}
 			\multicolumn{1}{|c|}{Offset Fault} & \multicolumn{5}{|c|} {82.33} & \multicolumn{1}{|c|}{SVM}\\
 			\cline{2-7}
 			\multicolumn{1}{|c|}{} & \multicolumn{5}{|c|} {79.12} & \multicolumn{1}{|c|}{CART}\\
 			\cline{2-7}
 			\multicolumn{1}{|c|}{}  & \multicolumn{5}{|c|} {79.12} & \multicolumn{1}{|c|}{RF}\\
 			\cline{1-7}
 			\multicolumn{1}{|c|}{} & \textbf{99.67} & \textbf{99.65} & \textbf{99.65} & \textbf{99.65} & \textbf{99.65} & \multicolumn{1}{|c|}{DSTE}\\
 			\cline{2-7}
 			\multicolumn{1}{|c|}{Gain Fault} & 81.97 & 81.60 & 81.50 & 81.47 & 81.45 & \multicolumn{1}{|c|}{SVM}\\
 			\cline{2-7}
 			\multicolumn{1}{|c|}{} & 99.60 & 99.60 & 99.60 & 99.60 & 99.60 & \multicolumn{1}{|c|}{CART}\\
 			\cline{2-7}
			\multicolumn{1}{|c|}{} & 99.60 & 99.60 & 99.60 & 99.60 & 99.60 & \multicolumn{1}{|c|}{RF}\\
 			\cline{1-7}
 			\multicolumn{1}{|c|}{} & \multicolumn{5}{|c|} {\textbf{99.62}} & \multicolumn{1}{|c|}{DSTE}\\
 			\cline{2-7}
 			\multicolumn{1}{|c|}{Out-of-Bounds Fault} & \multicolumn{5}{|c|} {82.33} & \multicolumn{1}{|c|}{SVM}\\
 			\cline{2-7}
 			\multicolumn{1}{|c|}{} & \multicolumn{5}{|c|} {99.60} & \multicolumn{1}{|c|}{CART}\\
 			\cline{2-7}
 			\multicolumn{1}{|c|}{}  & \multicolumn{5}{|c|} {99.60} & \multicolumn{1}{|c|}{RF}\\
 			\cline{1-7}
 			\multicolumn{1}{|c|} {}& \multicolumn{5}{|c|} {\textbf{99.62}} & \multicolumn{1}{|c|}{DSTE}\\
 			\cline{2-7}
 			\multicolumn{1}{|c|}{Data-Loss Fault} & \multicolumn{5}{|c|} {83.7} & \multicolumn{1}{|c|}{SVM}\\
 			\cline{2-7}
 			\multicolumn{1}{|c|}{} & \multicolumn{5}{|c|} {99.60} & \multicolumn{1}{|c|}{CART}\\
 			\cline{2-7}
 			\multicolumn{1}{|c|}{}  & \multicolumn{5}{|c|} {99.60} & \multicolumn{1}{|c|}{RF}\\
 			\cline{1-7}
 			\multicolumn{1}{|c|} {Fault Rate} & \multicolumn{5}{|c|} {\textbf{30\%}} &\\
 			\cline{1-6}
 			$\beta$ & 2 & 4 & 6 & 8 & 10\\
 			\cline{1-7}
 			\multicolumn{1}{|c|} {}& \multicolumn{5}{|c|} {\textbf{98.04}} & \multicolumn{1}{|c|}{DSTE}\\
 			\cline{2-7}
 			\multicolumn{1}{|c|}{Offset Fault} & \multicolumn{5}{|c|} {84.34} & \multicolumn{1}{|c|}{SVM}\\
 			\cline{2-7}
 			\multicolumn{1}{|c|}{} & \multicolumn{5}{|c|} {83.52} & \multicolumn{1}{|c|}{CART}\\
 			\cline{2-7}
 			\multicolumn{1}{|c|}{}  & \multicolumn{5}{|c|} {83.06} & \multicolumn{1}{|c|}{RF}\\
 			\cline{1-7}
 			\multicolumn{1}{|c|}{} & \textbf{99.70} & \textbf{99.68} & \textbf{99.68} & \textbf{99.67} & \textbf{99.67} & \multicolumn{1}{|c|}{DSTE}\\
 			\cline{2-7}
 			\multicolumn{1}{|c|}{Gain Fault} & 84.06 & 83.74 & 83.63 & 83.61 & 83.58 & \multicolumn{1}{|c|}{SVM}\\
 			\cline{2-7}
 			\multicolumn{1}{|c|}{} & 99.30 & 99.30 & 99.30 & 99.30 & 99.30 & \multicolumn{1}{|c|}{CART}\\
 			\cline{2-7}
 			\multicolumn{1}{|c|}{} & 99.23 & 99.40 & 99.23 & 99.11 & 99.11 & \multicolumn{1}{|c|}{RF}\\
 			\cline{1-7}
 			\multicolumn{1}{|c|}{} & \multicolumn{5}{|c|} {\textbf{99.69}} & \multicolumn{1}{|c|}{DSTE}\\
 			\cline{2-7}
 			\multicolumn{1}{|c|}{Out-of-Bounds Fault} & \multicolumn{5}{|c|} {84.35} & \multicolumn{1}{|c|}{SVM}\\
 			\cline{2-7}
 			\multicolumn{1}{|c|}{} & \multicolumn{5}{|c|} {99.27} & \multicolumn{1}{|c|}{CART}\\
 			\cline{2-7}
 			\multicolumn{1}{|c|}{}  & \multicolumn{5}{|c|} {99.07} & \multicolumn{1}{|c|}{RF}\\
 			\cline{1-7}
 			\multicolumn{1}{|c|}{} & \multicolumn{5}{|c|} {\textbf{99.70}} & \multicolumn{1}{|c|}{DSTE}\\
 			\cline{2-7}
 			\multicolumn{1}{|c|}{Data-Loss Fault} & \multicolumn{5}{|c|} {85.71} & \multicolumn{1}{|c|}{SVM}\\
 			\cline{2-7}
 			\multicolumn{1}{|c|}{} & \multicolumn{5}{|c|} {99.23} & \multicolumn{1}{|c|}{CART}\\
 			\cline{2-7}
 			\multicolumn{1}{|c|}{}  & \multicolumn{5}{|c|} {99.30} & \multicolumn{1}{|c|}{RF}\\
 			\cline{1-7}
 			 \multicolumn{1}{|c|} {Fault Rate} & \multicolumn{5}{|c|} {\textbf{40\%}} &\\
 			\cline{1-6}
 			$\beta$ & 2 & 4 & 6 & 8 & 10\\
 			\cline{1-7}
 			\multicolumn{1}{|c|} {}& \multicolumn{5}{|c|} {\textbf{99.44}} & \multicolumn{1}{|c|}{DSTE}\\
 			\cline{2-7}
 			\multicolumn{1}{|c|}{Offset Fault} & \multicolumn{5}{|c|} {85.9} & \multicolumn{1}{|c|}{SVM}\\
 			\cline{2-7}
 			\multicolumn{1}{|c|}{} & \multicolumn{5}{|c|} {82.5} & \multicolumn{1}{|c|}{CART}\\
 			\cline{2-7}
 			\multicolumn{1}{|c|}{}  & \multicolumn{5}{|c|} {83.87} & \multicolumn{1}{|c|}{RF}\\
 			\cline{1-7}
 			\multicolumn{1}{|c|}{} & \textbf{99.45} & \textbf{99.43} & \textbf{99.41} & \textbf{99.40} & \textbf{99.40}& \multicolumn{1}{|c|}{DSTE}\\
 			\cline{2-7}
 			\multicolumn{1}{|c|}{Gain Fault} & 85.66 & 85.35 & 85.34 & 85.27 & 85.25 & \multicolumn{1}{|c|}{SVM}\\
 			\cline{2-7}
 			\multicolumn{1}{|c|}{} & 98.34 & 98.34 & 98.34 & 98.34 & 98.34 & \multicolumn{1}{|c|}{CART}\\
 			\cline{2-7}
 			\multicolumn{1}{|c|}{} & 99.32 & 99.32 & 99.32 & 99.32 & 99.32 & \multicolumn{1}{|c|}{RF}\\
 			\cline{1-7}
 			\multicolumn{1}{|c|}{} & \multicolumn{5}{|c|} {\textbf{99.71}} & \multicolumn{1}{|c|}{DSTE}\\
 			\cline{2-7}
 			\multicolumn{1}{|c|}{Out-of-Bounds Fault} & \multicolumn{5}{|c|} {85.9} & \multicolumn{1}{|c|}{SVM}\\
 			\cline{2-7}
 			\multicolumn{1}{|c|}{} & \multicolumn{5}{|c|} {98.34} & \multicolumn{1}{|c|}{CART}\\
 			\cline{2-7}
 			\multicolumn{1}{|c|}{}  & \multicolumn{5}{|c|} {99.39} & \multicolumn{1}{|c|}{RF}\\
 			\cline{1-7}
 			\multicolumn{1}{|c|}{} & \multicolumn{5}{|c|} {\textbf{99.71}} & \multicolumn{1}{|c|}{DSTE}\\
 			\cline{2-7}
 			\multicolumn{1}{|c|}{Data-Loss Fault} & \multicolumn{5}{|c|} {89.05} & \multicolumn{1}{|c|}{SVM}\\
 			\cline{2-7}
 			\multicolumn{1}{|c|}{} & \multicolumn{5}{|c|} {98.34} & \multicolumn{1}{|c|}{CART}\\
 			\cline{2-7}
 			\multicolumn{1}{|c|}{}  & \multicolumn{5}{|c|} {99.39} & \multicolumn{1}{|c|}{RF}\\
 			\cline{1-7}
 			 \multicolumn{1}{|c|} {Fault Rate} & \multicolumn{5}{|c|} {\textbf{50\%}} &\\
 			\cline{1-6}
 			$\beta$ & 2 & 4 & 6 & 8 & 10\\
 			\cline{1-7}
 			\multicolumn{1}{|c|} {}& \multicolumn{5}{|c|} {\textbf{99.62}} & \multicolumn{1}{|c|}{DSTE}\\
 			\cline{2-7}
 			\multicolumn{1}{|c|}{Offset Fault} & \multicolumn{5}{|c|} {88.85} & \multicolumn{1}{|c|}{SVM}\\
 			\cline{2-7}
 			\multicolumn{1}{|c|}{} & \multicolumn{5}{|c|} {86.46} & \multicolumn{1}{|c|}{CART}\\
 			\cline{2-7}
 			\multicolumn{1}{|c|}{}  & \multicolumn{5}{|c|} {87.06} & \multicolumn{1}{|c|}{RF}\\
 			\cline{1-7}
 			\multicolumn{1}{|c|}{} & \textbf{99.78} & \textbf{99.75} & \textbf{99.75} & \textbf{99.75} & \textbf{99.75} & \multicolumn{1}{|c|}{DSTE}\\
 			\cline{2-7}
 			\multicolumn{1}{|c|}{Gain Fault} & 88.64 & 88.5 & 88.41 & 88.41 & 88.41 & \multicolumn{1}{|c|}{SVM}\\
 			\cline{2-7}
 			\multicolumn{1}{|c|}{} & 99.18 & 99.18 & 99.18 & 99.18 & 99.18 & \multicolumn{1}{|c|}{CART}\\
 			\cline{2-7}
 			\multicolumn{1}{|c|}{} & 99.62 & 99.62 & 99.62 & 99.62 & 99.62 & \multicolumn{1}{|c|}{RF}\\
 			\cline{1-7}
 			\multicolumn{1}{|c|}{} & \multicolumn{5}{|c|} {\textbf{99.78}} & \multicolumn{1}{|c|}{DSTE}\\
 			\cline{2-7}
 			\multicolumn{1}{|c|}{Out-of-Bounds Fault} & \multicolumn{5}{|c|} {88.84} & \multicolumn{1}{|c|}{SVM}\\
 			\cline{2-7}
 			\multicolumn{1}{|c|}{} & \multicolumn{5}{|c|} {99.18} & \multicolumn{1}{|c|}{CART}\\
 			\cline{2-7}
 			\multicolumn{1}{|c|}{}  & \multicolumn{5}{|c|} {99.59} & \multicolumn{1}{|c|}{RF}\\
 			\cline{1-7}
 			\multicolumn{1}{|c|}{} & \multicolumn{5}{|c|} {\textbf{99.84}} & \multicolumn{1}{|c|}{DSTE}\\
 			\cline{2-7}
 			\multicolumn{1}{|c|}{Data-Loss Fault} & \multicolumn{5}{|c|} {91.7} & \multicolumn{1}{|c|}{SVM}\\
 			\cline{2-7}
 			\multicolumn{1}{|c|}{} & \multicolumn{5}{|c|} {99.18} & \multicolumn{1}{|c|}{CART}\\
 			\cline{2-7}
 			\multicolumn{1}{|c|}{}  & \multicolumn{5}{|c|} {99.52} & \multicolumn{1}{|c|}{RF}\\
 			\cline{1-7}
 		\end{tabular}
 	}
 \end{table}
\subsubsection{Analysis of laboratory data set}
Experiments have also been conducted on laboratory data set for different kinds of fault detection to further validate the effectiveness of the proposed method. Table~\ref{tab_4} shows the performance comparison of the proposed method with the existing state-of-the-art classifiers. From the table it can be seen that DSTE performs competitively better than the compared algorithms with RF also showing quite good performance. 
	\begin{table}
	\setlength{\tabcolsep}{6.7pt}
	\vspace{-0.9em}    
	\caption{Result of Faulty Node Analysis on Experimental Data Set (Fault Rate = 40\%, $\alpha =10$, $\beta = 6$). The best values are given in bold.}
	\label{tab_4}
	\resizebox{1.02\textwidth}{!}{
		\begin{tabular}{c|c|c|c|c|c|c}
			\cline{2-6}

			\multicolumn{1}{c|} {} & Accuracy (\%) & FPR (\%) & Precision (\%) & Sensitivity (\%) & Specificity (\%) & \multicolumn{1}{|c} {}\\
			\cline{1-7}
			\multicolumn{1}{|c|} {} & \textbf{99.96} & \textbf{0} & \textbf{100} &  \textbf{100} & \textbf{100} & \multicolumn{1}{|c|}{DSTE}\\
			\multicolumn{1}{|c|} {Offset Fault} & 62.8 & \textbf{0} & \textbf{100} &  13.63 & \textbf{100 }& \multicolumn{1}{|c|}{SVM}\\
			\multicolumn{1}{|c|} {} &  99.91 & \textbf{0 }& 98.29 &  98.87 & \textbf{100} & \multicolumn{1}{|c|}{CART}\\
			\multicolumn{1}{|c|} {} &  99.94 & \textbf{0}& 99.19 &  99.65 & \textbf{100} & \multicolumn{1}{|c|}{RF}\\
			\cline{1-7}
			\multicolumn{1}{|c|} {} & \textbf{99.97} & \textbf{0} & \textbf{100} &  \textbf{100} & \textbf{100} & \multicolumn{1}{|c|}{DSTE}\\
			\multicolumn{1}{|c|} {Gain Fault} & 62.01 & 66.74 & 53.14 &  100 & 33.25 & \multicolumn{1}{|c|}{SVM}\\
			\multicolumn{1}{|c|} {} &  99.91 & \textbf{0} & 99.18 &  98.94 & 99.32 & \multicolumn{1}{|c|}{CART}\\
			\multicolumn{1}{|c|} {} & \textbf{99.97} & \textbf{0} & \textbf{100} &  \textbf{100} & \textbf{100} & \multicolumn{1}{|c|}{RF}\\
			\cline{1-7}
			\multicolumn{1}{|c|} {} & \textbf{99.98} & \textbf{0} & \textbf{100} &  \textbf{100} & \textbf{100} & \multicolumn{1}{|c|}{DSTE}\\
			\multicolumn{1}{|c|} {Out-of-Bounds} & 62.97 & \textbf{0} & \textbf{100} &  13.63 & \textbf{100} & \multicolumn{1}{|c|}{SVM}\\
			\multicolumn{1}{|c|} {Fault} &  99.91 & \textbf{0} & 99.18 &  98.94 & \textbf{100} & \multicolumn{1}{|c|}{CART}\\
			\multicolumn{1}{|c|} {} &  99.96 & \textbf{0} & \textbf{100} & \textbf{100} & \textbf{100} & \multicolumn{1}{|c|}{RF}\\
			\cline{1-7}
			\multicolumn{1}{|c|} {} & \textbf{99.98} &\textbf{0} & \textbf{100} &  \textbf{100} & \textbf{100} & \multicolumn{1}{|c|}{DSTE}\\
			\multicolumn{1}{|c|} {Data Fault} & 99.21 & \textbf{0} & \textbf{100} &  98.18 & \textbf{100} & \multicolumn{1}{|c|}{SVM}\\
			\multicolumn{1}{|c|} {} &  99.34 & \textbf{0} & \textbf{100}&  98.48 & \textbf{100} & \multicolumn{1}{|c|}{CART}\\
			\multicolumn{1}{|c|} {} &  99.96 & \textbf{0} &\textbf{100} & \textbf{100} & \textbf{100} & \multicolumn{1}{|c|}{RF}\\
			\cline{1-7}
		\end{tabular}
	}
	\end{table}

\begin{figure}[H]
		\centering
		\includegraphics[width=7.2cm,height=5.2cm]{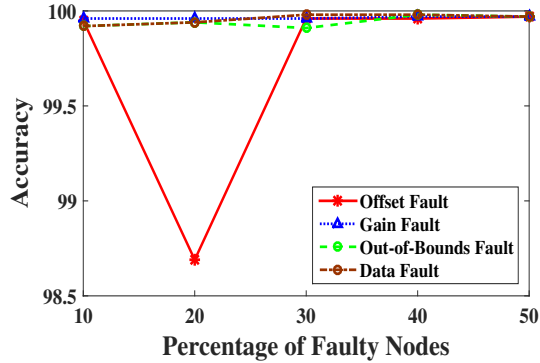}
	\caption{Accuracy of DSTE for different types of faults}\label{accplot}
\end{figure}

Fig.~\ref{accplot} shows the accuracy of the proposed method for different kinds of faults for different percentage of faulty nodes at $\beta=2$. As can be seen from the figure, DSTE shows quite high accuracy while detecting all kinds of faults.

The details of the accuracy values for each of the faults for all combination of features or parameters are depicted in Table~\ref{tab_5}. 
	\begin{table}
	\setlength{\tabcolsep}{5.5pt}
	\vspace{-0.9em}    
	\caption{Performance Evaluation of DSTE in Terms of Accuracy (\%) for Different Kinds of Faults (Fault Rate = 10\%, $\alpha =10$, $\beta = 2$) }
	\label{tab_5}
	\resizebox{1\textwidth}{!}{
		\begin{tabular}{p{2.5cm}l l l l}
			\cline{1-5}

			Parameters & Offset Fault & Gain Fault & Out-of-Bounds  & Data Fault\\
			& & & Fault &\\
			\cline{1-5}
			T & 99.25 & 99.23 & 99.44 & 99.19 \\
			H & 88.12 & 99.17 & 99.11 & 99.43 \\
			L & 97.21 & 98.95 & 96.56 & 98.82\\
			T+H & 99.13 & 99.56 & 99.21 & 99.35\\
			H+L & 90.2 & 99.18 & 99.11 & 99.53\\
			T+L & 99.65 & 99.37 & 99.46 & 99.24\\
			T+H+L & 99.92 & 99.90 & 99.96 & 99.97\\
			\cline{1-5}
			\multicolumn{5}{l} {T = Temperature ($^0$C), H = Humidity (\%), L = Light (Lux)}\\
		\end{tabular}
	}
\end{table}
From the table, it is quite evident that DSTE produces quite satisfactory result for all kinds of faults. The fault detection improves when the number of fused parameters increase, thereby emphasising the importance of Dempster's Combination rule. 
\section{Conclusion}
\label{con}
IoT-based devices which are equipped with the tool to detect faulty nodes are very much the need of the hour. This work has adopted Dempster-Shafer Theory of Evidence to fuse data from different sensors to conclude about the status of a sensor node. This work is driven by the fact that fusing data from different data sources is always a better strategy than to derive conclusion form a single source. In this work, a mass assignment function is put forward which uses the concept of normal data distribution. At first the data is trained with the proposed method and then it is applied on the test cases. The proposed method is verified with a benchmark data set as well as on a real-life data set taken from laboratory set-up. Different types of faults were incorporated into both the data sets to prepare the final version of the data sets for conducting the analysis. Based on the experimental analysis, it can be concluded that the proposed method has a very high accuracy (99.8\% and 99.9\% for labelled and laboratory data set respectively) and also shows overall superior performance when compared to other existing machine learning techniques.

As a future work, the authors would like to create a hybrid classifier which will be an amalgamation of DSTE and other learning theories to come up with a method for detection of faulty nodes in IoT-based environment. 
\bibliographystyle{plain}
\bibliography{Reference} 

\begin{thebibliography}{10}

\bibitem{Aabid}
A.~{Abid}, A.~{Kachouri}, and A.~{Mahfoudhi}.
\newblock Outlier detection for wireless sensor networks using density-based
  clustering approach.
\newblock {\em IET Wireless Sensor Systems}, 7(4):83--90, 2017.

\bibitem{Aggarwal2015}
Charu~C. Aggarwal.
\newblock {\em Outlier Analysis}, pages 237--263.
\newblock Springer International Publishing, 2015.

\bibitem{JAn}
J.~{An}, M.~{Hu}, L.~{Fu}, and J.~{Zhan}.
\newblock A novel fuzzy approach for combining uncertain conflict evidences in
  the dempster-shafer theory.
\newblock {\em IEEE Access}, 7:7481--7501, 2019.

\bibitem{Dempster2008}
Arthur~P. Dempster.
\newblock {\em A Generalization of Bayesian Inference}, pages 73--104.
\newblock Springer Berlin Heidelberg, Berlin, Heidelberg, 2008.

\bibitem{Djenouri}
Y.~{Djenouri}, A.~{Belhadi}, J.~C. {Lin}, and A.~{Cano}.
\newblock Adapted k-nearest neighbors for detecting anomalies on
  spatio–temporal traffic flow.
\newblock {\em IEEE Access}, 7:10015--10027, 2019.

\bibitem{FAWZY}
Asmaa Fawzy, Hoda~M.O. Mokhtar, and Osman Hegazy.
\newblock Outliers detection and classification in wireless sensor networks.
\newblock {\em Egyptian Informatics Journal}, 14(2):157 -- 164, 2013.

\bibitem{Ghorbel}
O.~{Ghorbel}, W.~{Ayedi}, H.~{Snoussi}, and M.~{Abid}.
\newblock Fast and efficient outlier detection method in wireless sensor
  networks.
\newblock {\em IEEE Sensors Journal}, 15(6):3403--3411, June 2015.

\bibitem{HAJIHEIDARI2019}
Somayye Hajiheidari, Karzan Wakil, Maryam Badri, and Nima~Jafari Navimipour.
\newblock Intrusion detection systems in the {I}nternet of things: A
  comprehensive investigation.
\newblock {\em Computer Networks}, 2019.

\bibitem{hawkins}
Douglas~M Hawkins.
\newblock {\em Identification of outliers}, volume~11.
\newblock Springer, 1980.

\bibitem{LJin}
L.~{Jin}, J.~{Chen}, and X.~{Zhang}.
\newblock An outlier fuzzy detection method using fuzzy set theory.
\newblock {\em IEEE Access}, 7:59321--59332, 2019.

\bibitem{Wlu}
W.~{Lu}, Y.~{Cheng}, C.~{Xiao}, S.~{Chang}, S.~{Huang}, B.~{Liang}, and
  T.~{Huang}.
\newblock Unsupervised sequential outlier detection with deep architectures.
\newblock {\em IEEE Transactions on Image Processing}, 26(9):4321--4330, Sep.
  2017.

\bibitem{MUHAMMED2017}
Thaha Muhammed and Riaz~Ahmed Shaikh.
\newblock An analysis of fault detection strategies in wireless sensor
  networks.
\newblock {\em Journal of Network and Computer Applications}, 78:267 -- 287,
  2017.

\bibitem{BQin}
B.~{Qin} and F.~{Xiao}.
\newblock A non-parametric method to determine basic probability assignment
  based on kernel density estimation.
\newblock {\em IEEE Access}, 6:73509--73519, 2018.

\bibitem{Reppa}
V.~{Reppa}, P.~{Papadopoulos}, M.~M. {Polycarpou}, and C.~G. {Panayiotou}.
\newblock A distributed architecture for hvac sensor fault detection and
  isolation.
\newblock {\em IEEE Transactions on Control Systems Technology},
  23(4):1323--1337, July 2015.

\bibitem{shafer197}
Glenn Shafer.
\newblock {\em A mathematical theory of evidence}, volume~42.
\newblock Princeton university press, 1976.

\bibitem{shepherd2003}
J~Marshall Shepherd and Steven~J Burian.
\newblock Detection of urban-induced rainfall anomalies in a major coastal
  city.
\newblock {\em Earth {I}nteractions}, 7(4):1--17, 2003.

\bibitem{Suthaharan}
S.~{Suthaharan}, M.~{Alzahrani}, S.~{Rajasegarar}, C.~{Leckie}, and
  M.~{Palaniswami}.
\newblock Labelled data collection for anomaly detection in wireless sensor
  networks.
\newblock In {\em 2010 Sixth International Conference on Intelligent Sensors,
  Sensor Networks and Information Processing}, pages 269--274, Dec 2010.

\bibitem{Warriach}
E.~U. {Warriach} and K.~{Tei}.
\newblock Fault detection in wireless sensor networks: A machine learning
  approach.
\newblock In {\em 2013 IEEE 16th International Conference on Computational
  Science and Engineering}, pages 758--765, Dec 2013.

\bibitem{Xia}
J.~{Xia}, Y.~{Feng}, L.~{Liu}, D.~{Liu}, and L.~{Fei}.
\newblock An evidential reliability indicator-based fusion rule for
  dempster-shafer theory and its applications in classification.
\newblock {\em IEEE Access}, 6:24912--24924, 2018.

\bibitem{Xie}
K.~{Xie}, X.~{Li}, X.~{Wang}, J.~{Cao}, G.~{Xie}, J.~{Wen}, D.~{Zhang}, and
  Z.~{Qin}.
\newblock On-line anomaly detection with high accuracy.
\newblock {\em IEEE/ACM Transactions on Networking}, 26(3):1222--1235, June
  2018.

\bibitem{Yessembayev}
A.~{Yessembayev}, D.~{Sarkar}, and F.~{Sikder}.
\newblock Detection of good and bad sensor nodes in the presence of malicious
  attacks and its application to data aggregation.
\newblock {\em IEEE Transactions on Signal and Information Processing over
  Networks}, 4(3):549--563, Sep. 2018.

\bibitem{Tyu}
T.~{Yu}, X.~{Wang}, and A.~{Shami}.
\newblock Recursive principal component analysis-based data outlier detection
  and sensor data aggregation in iot systems.
\newblock {\em IEEE Internet of Things Journal}, 4(6):2207--2216, Dec 2017.

\bibitem{ZERVAS}
E.~Zervas, A.~Mpimpoudis, C.~Anagnostopoulos, O.~Sekkas, and
  S.~Hadjiefthymiades.
\newblock Multisensor data fusion for fire detection.
\newblock {\em Information Fusion}, 12(3):150 -- 159, 2011.

\bibitem{HZhang}
H.~{Zhang}, Q.~{Zhang}, J.~{Liu}, and H.~{Guo}.
\newblock Fault detection and repairing for intelligent connected vehicles
  based on dynamic bayesian network model.
\newblock {\em IEEE Internet of Things Journal}, 5(4):2431--2440, Aug 2018.

\bibitem{HZhu}
H.~{Zhu}, S.~{Feng}, and F.~{Yu}.
\newblock Parking detection method based on finite-state machine and
  collaborative decision-making.
\newblock {\em IEEE Sensors Journal}, 18(23):9829--9839, Dec 2018.

\bibitem{zidi2017}
Salah Zidi, Tarek Moulahi, and Bechir Alaya.
\newblock Fault detection in wireless sensor networks through svm classifier.
\newblock {\em IEEE Sensors Journal}, 18(1):340--347, 2017.

\end{thebibliography}

\end{document}